\documentclass{article}
\usepackage{amssymb}
\usepackage{amsmath}
\usepackage{amsfonts}
\usepackage{booktabs}
\usepackage{enumitem}
\usepackage{multirow}
\usepackage{mathtools}
\usepackage[font=normal]{caption}
\setlist[enumerate]{noitemsep, topsep=0pt}
\usepackage[ruled,vlined]{algorithm2e}
\usepackage{float}
\usepackage{listings}
\usepackage{xcolor} 
\usepackage{graphicx}
\usepackage[toc,page,header]{appendix}
\usepackage{minitoc}
\usepackage{siunitx} 
\graphicspath{ {./figures/} }

\usepackage{color, soul} 

\newcommand{\sect}[1]{Sec.~\ref{#1}}

\newcommand{\grtm}{GRT}

\DeclareMathOperator*{\argmin}{arg\,min}

\usepackage[preprint]{corl_2025} 

\title{Geometric Red-Teaming for Robotic Manipulation}

\author{
  Divyam~Goel\textsuperscript{1} \quad
  Yufei~Wang\textsuperscript{1} \quad
  Tiancheng~Wu\textsuperscript{1} \quad
  Guixiu~Qiao\textsuperscript{2} \quad
  Pavel~Piliptchak\textsuperscript{2} \And
  David~Held\textsuperscript{1}\thanks{Equal advising} \quad
  Zackory~Erickson\textsuperscript{1}\footnotemark[1] \\
  \textsuperscript{1}Robotics Institute, Carnegie Mellon University, Pittsburgh, PA, USA \\
  \texttt{\{divyamg,yufeiw2,tianche3,dheld,zerickso\}@andrew.cmu.edu} \\
  \textsuperscript{2}National Institute of Standards and Technology, Gaithersburg, MD, USA \\
  \texttt{\{guixiu.qiao,pavel.piliptchak\}@nist.gov} \\
}

\begin{document}
\maketitle



\begin{abstract}
    Standard evaluation protocols in robotic manipulation typically assess policy performance over curated, in-distribution test sets, offering limited insight into how systems fail under plausible variation. 
    We introduce Geometric Red-Teaming (\grtm), a red-teaming framework that probes robustness through object-centric geometric perturbations, automatically generating CrashShapes---structurally valid, user-constrained mesh deformations that trigger catastrophic failures in pre-trained manipulation policies. 
    The method integrates a Jacobian field–based deformation model with a gradient-free, simulator-in-the-loop optimization strategy.
    Across insertion, articulation, and grasping tasks, \grtm~consistently discovers deformations that collapse policy performance, revealing brittle failure modes missed by static benchmarks. 
    By combining task-level policy rollouts with constraint-aware shape exploration, we aim to build a general purpose framework for structured, object-centric robustness evaluation in robotic manipulation.
    We additionally show that fine-tuning on individual CrashShapes, a process we refer to as blue-teaming, improves task success by up to 60 percentage points on those shapes, while preserving performance on the original object, demonstrating the utility of red-teamed geometries for targeted policy refinement.
    Finally, we validate both red-teaming and blue-teaming results with a real robotic arm, observing that simulated CrashShapes reduce task success from 90\% to as low as 22.5\%, and that blue-teaming recovers performance to up to 90\% on the corresponding real-world geometry---closely matching simulation outcomes.
    Videos and code can be found on our project website: \url{https://georedteam.github.io/}.
\end{abstract}



\section{Introduction}
\label{sec:introduction}
\vspace{-2mm}

    Standard evaluation protocols in robotic manipulation often benchmark policies on curated, in-distribution test sets, providing limited insight into failure modes under plausible variation.
    Such evaluations often obscure vulnerabilities arising from subtle shifts in object geometry, which can unpredictably alter affordances, disrupt contact dynamics, and precipitate task failure.
    While adjacent fields like vision and language have developed systematic tools for probing model robustness under controlled input variations~\cite{hong2024curiosity,ma2023evolving,jankowski2024red}, analogous methods are only beginning to emerge in robotic manipulation~\cite{majumdar2025predictive,sagarmystery,duan2024aha,karnik2024embodied}.
    Specifically, no formal frameworks exist for systematically evaluating policy performance under plausible, task-relevant geometric perturbations, despite the centrality of object shape to manipulation.

    In this work, we pose the following question: \textit{Can we automatically discover failure-inducing object geometries, treating the policy strictly as a black box?} 
    To address this question, we cast the task as a \textit{red-teaming} problem, inspired by cybersecurity frameworks that proactively discover vulnerabilities via realistic and targeted stress tests.
    Our objective is to generate geometric deformations that induce catastrophic policy failures---deformed objects which we refer to as \textit{CrashShapes}---while enforcing a geometric prior governing permissible shape variations, ensuring that generated objects remain physically plausible and semantically coherent. 
    Since failure depends on how an object interacts with the policy and task environment, we rely on embodied simulation rollouts to reveal small shape perturbations that break the learned assumptions underlying grasp affordances, contact transitions, or control trajectories.
    We refer to this system as \grtm~
    (``Geometric Red-Teaming").

    Operationalizing this concept poses two primary challenges:
     (1) policy performance must be evaluated via simulator rollouts, which are non-differentiable and therefore preclude the use of gradient-based optimization methods to generate CrashShapes;
     (2) the deformation space is high-dimensional and must be explored under structural constraints that preserve mesh integrity and ensure physically plausible, semantically coherent variations.
    To address these, we propose \grtm, a modular framework integrating a physically-grounded deformation model with gradient-free optimization. 
    Our method defines a deformation via handle points (vertices actively displaced) and anchor points (fixed vertices), using a Jacobian-field formulation derived from the APAP mesh editing framework~\cite{yoo2024plausible}. 
    Both handle and anchor points can be manually specified or automatically selected via a vision-language model (VLM), facilitating flexible, user-guided, or task-conditioned perturbations. 
    The optimizer operates exclusively through simulator feedback, accommodating both binary and continuous success metrics, and generalizing across diverse object types and manipulation tasks.
    Finally, we propose a constrained variant of our method that enforces a bound on the average handle displacement, restricting the search to small deviations from the nominal geometry that still yield large performance degradations.

    \begin{figure}[!t]
        \centering
        \includegraphics[width=0.95\textwidth]{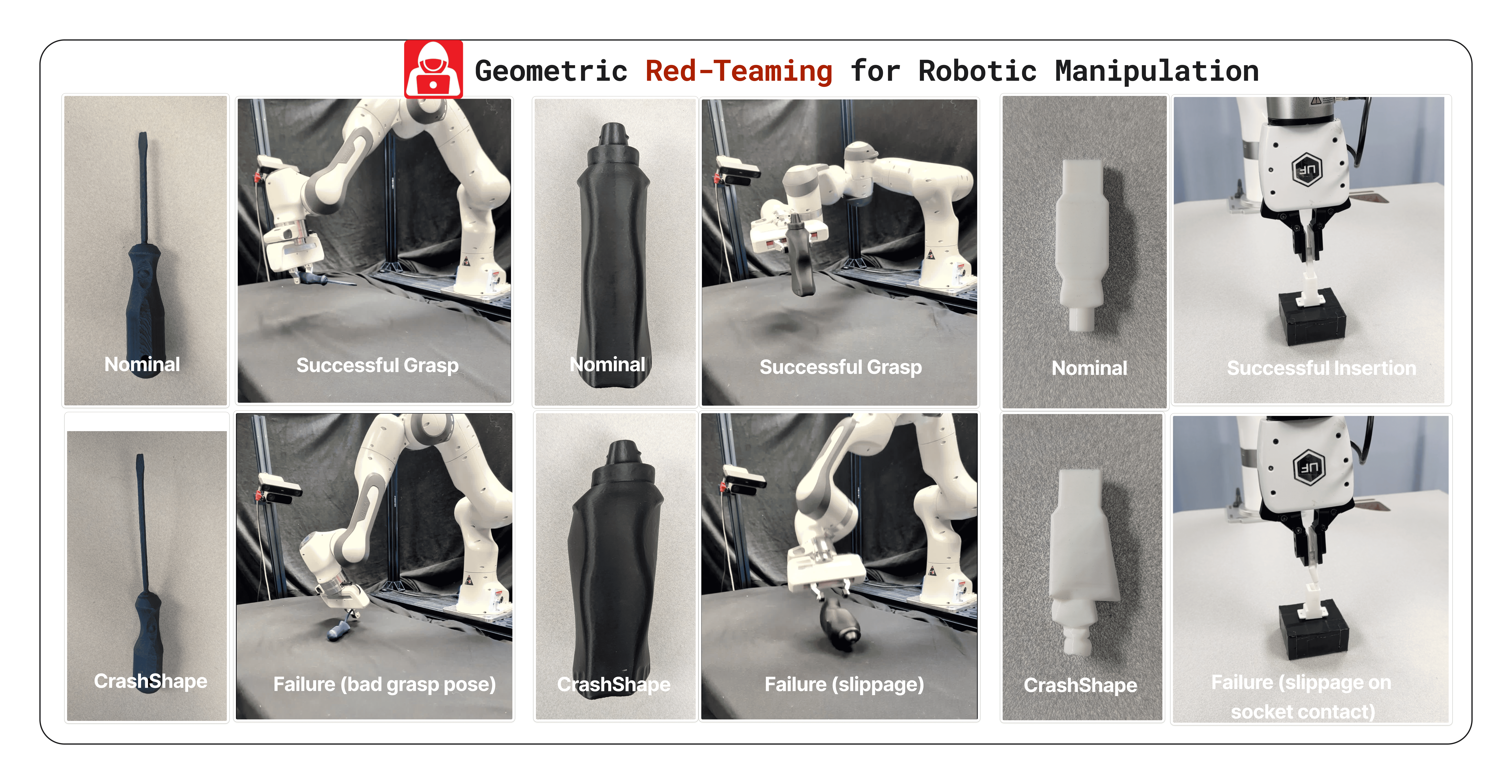}
        \caption{\textbf{GRT surfaces policy failures on a real robot from minimal, plausible geometry edits.} Top: nominal screwdriver, bottle, and USB plug succeed. Bottom: \emph{CrashShapes} induce bad grasp pose, grasp slippage, and insertion failure via in-gripper plug rotation at socket contact. Small, realistic deformations collapse policies that succeed on the original object.}
    
        \label{fig:pull_figure}
        \vspace{-5mm}
    \end{figure}

    This work presents the first red-teaming framework explicitly exploring 3D geometric deformation for robotic manipulation. 
    Existing methods target symbolic parameters~\cite{sagarmystery}, language instructions~\cite{karnik2024embodied}, or scene-level failure taxonomies~\cite{duan2024aha}. 
    Object geometry is another critical axis of failure. 
    We explicitly target object-centric manipulation policies by perturbing the 3D meshes that define both sensory input and interaction dynamics in simulation.
    
    In summary, this paper makes the following contributions:
        \begin{enumerate}
            \item We introduce \grtm, a policy-agnostic, simulator-in-the-loop framework that automatically discovers physically plausible CrashShapes inducing catastrophic ($>$ 50\%) failures in pre-trained manipulation policies.

            \item We validate \grtm~in simulation across three domains---high-precision industrial insertion, articulated drawer manipulation, and rigid-object grasping---demonstrating reliable failure discovery in each.

            \item We demonstrate the practical utility of our framework by showing that discovered CrashShapes transfer to a physical robot and that simple PPO fine-tuning recovers up to 60 percentage points of performance without degrading performance on nominal shapes.
        \end{enumerate}


\section{Related Work}
\label{sec:related_work}

    \paragraph{Evaluation Strategies in ML and Robotics}
    Evaluation in machine learning and robotics has long relied on benchmarks over canonical scenes or narrowly drawn object distributions~\cite{james2020rlbench,zhu2020robosuite,yu2020meta}, which obscure policy behavior under unseen, realistic variations of the task environment.
    In contrast, vision and language domains now routinely employ adversarial testing~\cite{kumar2023certifying,liu2024groot} and red-teaming strategies~\cite{tsai2023ring,hong2024curiosity,ma2023evolving,jankowski2024red} to reveal behavioral blind spots.
    Despite the increasing autonomy and deployment of robotic systems, comparable dynamic evaluation tools tailored to manipulation remain limited. 
    \grtm~addresses this gap through systematic, object-centric geometric perturbations.

    \vspace{-2.5mm}
    \paragraph{Evaluating Policies under Geometric Variation}
    Object geometry presents a critical axis for generalization in robotic manipulation.
    DoorGym~\cite{urakami2019doorgym} and ManiSkill~\cite{mu2021maniskill} introduce procedural object variation within task families, while EGAD~\cite{morrison2020egad} adopts evolutionary strategies to generate datasets covering a spectrum of geometric complexity and grasp difficulty. 
    However, these datasets remain static, task-specific, and entirely agnostic to policy behavior.
    In contrast, we formulate geometric deformation as a continuous, policy-conditioned search for failure-inducing perturbations.

    \vspace{-2.5mm}
    \paragraph{Failure Mode Discovery for Robot Policies}
    Recent efforts in robustness evaluation have moved beyond static benchmarks toward active vulnerability discovery~\cite{sagarmystery,duan2024aha,karnik2024embodied,majumdar2025predictive}.
    RoboMD~\cite{sagarmystery}, for instance, employs RL to perturb scene attributes during policy rollouts, rewarding configurations that induce failure. 
    However, this methodology typically requires explicit parameterization of scene attributes, offering limited insight into failure modes rooted in fine-grained, continuous geometry.
    Similarly, AHA~\cite{duan2024aha} relies on predefined error taxonomies, constraining discovered failure types, while language-conditioned methods~\cite{karnik2024embodied} study semantic shifts but stay detached from object-physical interactions.
    In contrast, by applying physically plausible mesh deformations and simulator rollouts to surface geometric misgeneralizations, our method introduces 3D object geometry as a core axis of failure analysis.


\vspace{-2mm}
\section{Background}
\label{sec:background}
\vspace{-2mm}

    \paragraph{Mesh Deformation with Jacobian Fields}
        To generate structurally coherent deformations of 3D meshes, we adopt the first stage of APAP~\cite{yoo2024plausible}.
        Given a source mesh $M_0 = (V_0, F_0)$, where $V_0 \in \mathbb{R}^{V\times3}$ are vertex positions and $F_0$ denotes triangular faces, a local affine Jacobian $J_f \in \mathbb{R}^{3\times3}$ is assigned to each face $f \in F_0$.
        Deformation proceeds by specifying a set of handle points $H \subseteq V_0$ with target positions $T_h \in \mathbb{R}^{H\times3}$, and fixed anchor points $A\subseteq V_0$ with targets $T_a \in \mathbb{R}^{A\times3}$, preventing trivial global translations and allowing more control over deformations in local regions of the mesh.
        The deformed mesh $V^{*}$ is obtained by solving:
        \begin{equation}
        \footnotesize
            V^{*} = \argmin_V || LV - \nabla^T AJ ||^{2} + \lambda||K_aV - T_a||^2,
        \end{equation}
        where L is the cotangent Laplacian, $\nabla$ is the stacked per-face gradient operator, A is a mass matrix, J is the given Jacobian field, $K_a$ is an indicator matrix which selects anchor vertices, and $\lambda$ controls the constraint strength.
        This linear system is solved via a differentiable Cholesky solver. 
        Soft handle constraints are then imposed by minimizing the loss $\mathcal{L}_h = || K_hV^* - T_h ||^2$ through gradient descent on the underlying Jacobian field, where $K_h$ is an indicator matrix to select handle vertices.
        We omit the second-stage 2D diffusion prior used in~\citet{yoo2024plausible} due to its limited benefit in task-specific object domains and substantial computational overhead.
        For more details, see Appendix~\ref{app:apap_prior}.

    \vspace{-2.5mm}
    \paragraph{Gradient-Free Optimizers for Black-Box Search}
        We approach deformation parameter search as a black-box optimization problem, in which gradients with respect to policy performance are unavailable and evaluations are carried out via simulator rollouts. 
        Population-based, gradient-free methods such as CEM~\cite{de2005tutorial} and CMA-ES~\cite{hansen2006cma} are commonly used in such settings.
        \grtm~builds on TOPDM~\cite{charlesworth2021solving}, which introduces a selective perturbation strategy---at each iteration, only a random subset of parameters is modified, rather than perturbing the entire candidate vector. 
        This selective perturbation explores local refinements without overwriting globally effective structure, enabling sample-efficient discovery of subtle, failure-inducing deformations.


\vspace{-2mm}
\section{Problem Formulation}
\label{sec:problem_formulation}
\vspace{-2mm}

    We consider the problem of evaluating the geometric robustness of robotic manipulation policies by discovering physically plausible deformations of 3D object meshes that induce policy failure. 

    Let $\pi: \mathcal{S} \rightarrow \mathcal{A}$ denote a pre-trained manipulation policy, where $\mathcal{S}$ is the policy's observation space and $\mathcal{A}$ is the action space. 
    Each object instance is represented as a watertight triangle mesh $M = (V, F)$, with vertices $V \in \mathbb{R}^{n \times 3}$ and triangular faces $F$.
    A parameterized deformation operator $D_\theta: \mathbb{R}^{n \times 3} \rightarrow \mathbb{R}^{n \times 3}$, with $\theta \in \Theta$ representing the deformation parameters to be optimized, maps the original mesh to a deformed variant $\tilde{M} = (D_\theta(V), F)$.
    Given a task-specific success metric $\mathcal{J}(\pi, \tilde{M}) \in \mathbb{R}$ computed via simulation rollouts, we aim to discover deformation parameters that minimize task performance:
        \begin{equation}
        \footnotesize
            \theta^* = \argmin_{\theta \in \Theta, \; D_\theta(M) \in \mathcal{G}(M)} \mathcal{J}(\pi, D_\theta(M)),
        \end{equation}
    where $\mathcal{G}(M)$ denotes the set of physically plausible deformations of the nominal mesh $M$, ensuring that the search is restricted to task-relevant perturbations, rather than degenerate geometries.
    Optionally, we restrict search to constraining the average handle displacement.
    See Appendix~\ref{app:smoothness-score} for more details on this constrained variant of our method.

    \vspace{-2mm}
    \paragraph{Assumptions}
    We assume access to a pre-trained manipulation policy $\pi$.
    Each object is represented as a watertight, manifold triangle mesh $M = (V, F)$.
    We consider canonical objects on which the policy achieves high success, as established by training performance or empirical evaluation.
    We also assume access to a physics-based simulator capable of loading deformed meshes $\tilde{M} = D_\theta(M)$, executing $\pi$, and reporting a scalar task performance metric $\mathcal{J}(\pi, \tilde{M})$.
    The simulator must model object contact, dynamics, and task-environment interactions to yield meaningful evaluation signals. 
    Neither the simulator nor the task performance metric is assumed to be differentiable.


\begin{figure}[!t]
    \centering
    \includegraphics[width=\linewidth]{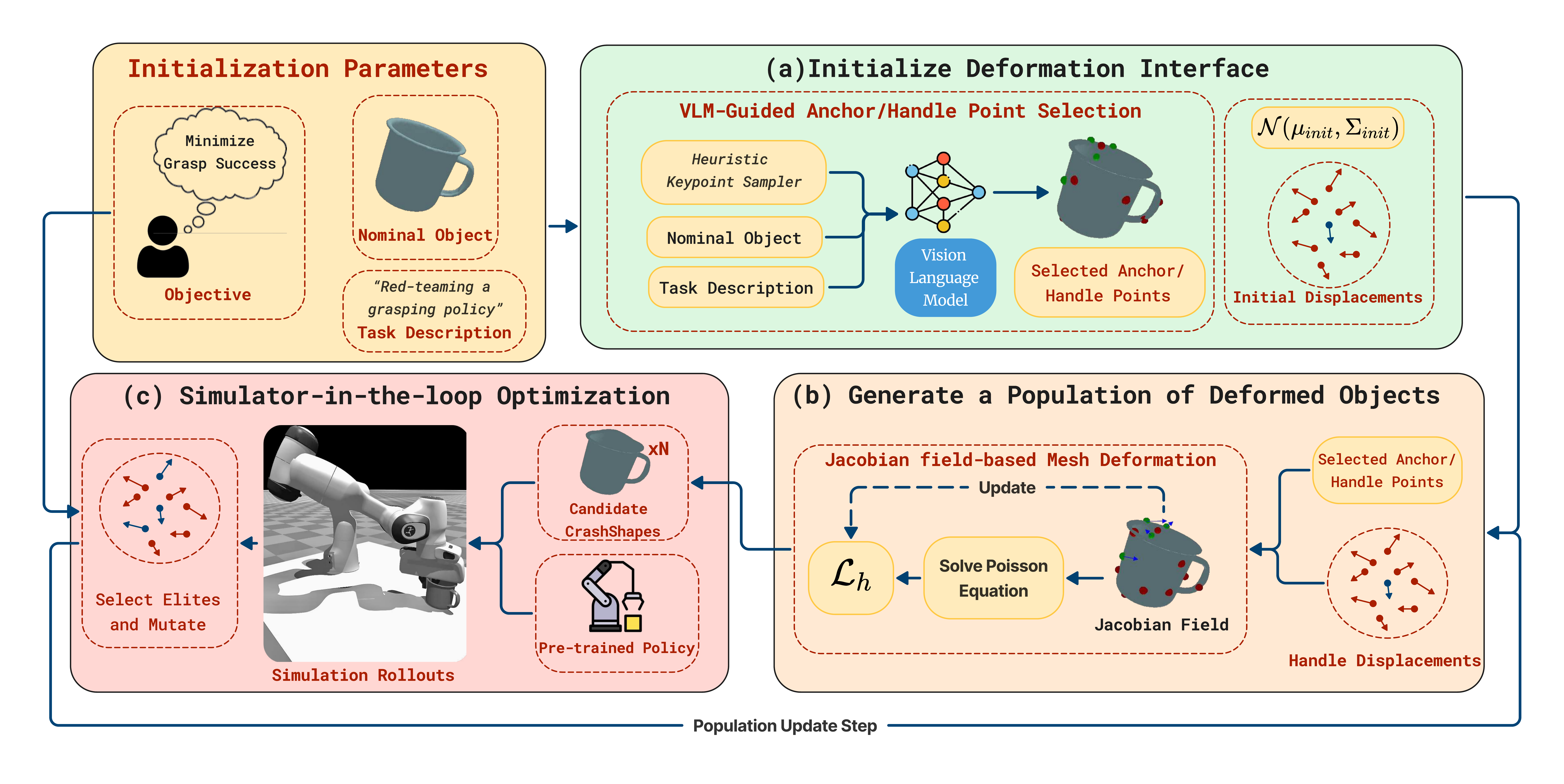}
    \vspace{-8mm}
    \caption{\textbf{System overview of \grtm.} Given a task description and nominal object (\emph{Initialization Parameters}), anchor and handle points are selected using a vision-language model (\textbf{a}). Handle displacements are sampled to define a population of deformation candidates. Each sample is converted into a perturbed mesh via Jacobian field-based optimization (\textbf{b}) and evaluated in simulation with a frozen policy (\textbf{c}). Deformations that induce failure are sampled to guide the next population.}
    \label{fig:system_figure}
    \vspace{-5mm}
\end{figure}

\vspace{-2.5mm}
\section{Method}
\label{sec:method}
\vspace{-2mm}
    
    We propose \grtm, an object-centric red-teaming framework for robotic manipulation policies, which defines a continuous search space over object geometries, targeting minimal perturbations that induce policy failures.
    \grtm~(see Figure~\ref{fig:system_figure}) consists of four key components.
    First, we expose a deformation interface on $M$ by selecting task-relevant handle and anchor points. 
    Second, we apply a Jacobian field-based deformation model to generate smooth perturbations of the object from handle displacements.
    Third, we evaluate $\pi$ on the deformed object $D_\theta(M)$ through simulation rollouts.
    Finally, we leverage gradient-free optimization to search over the deformation parameter space $\theta$, to discover physically plausible deformations that maximally impair policy performance.

    \vspace{-2mm}
    \subsection{VLM-Guided Mesh Deformation}
    \label{sec:vlm_mesh_deformation}
\begin{figure}[!t]
            \centering
            \includegraphics[width=\linewidth]{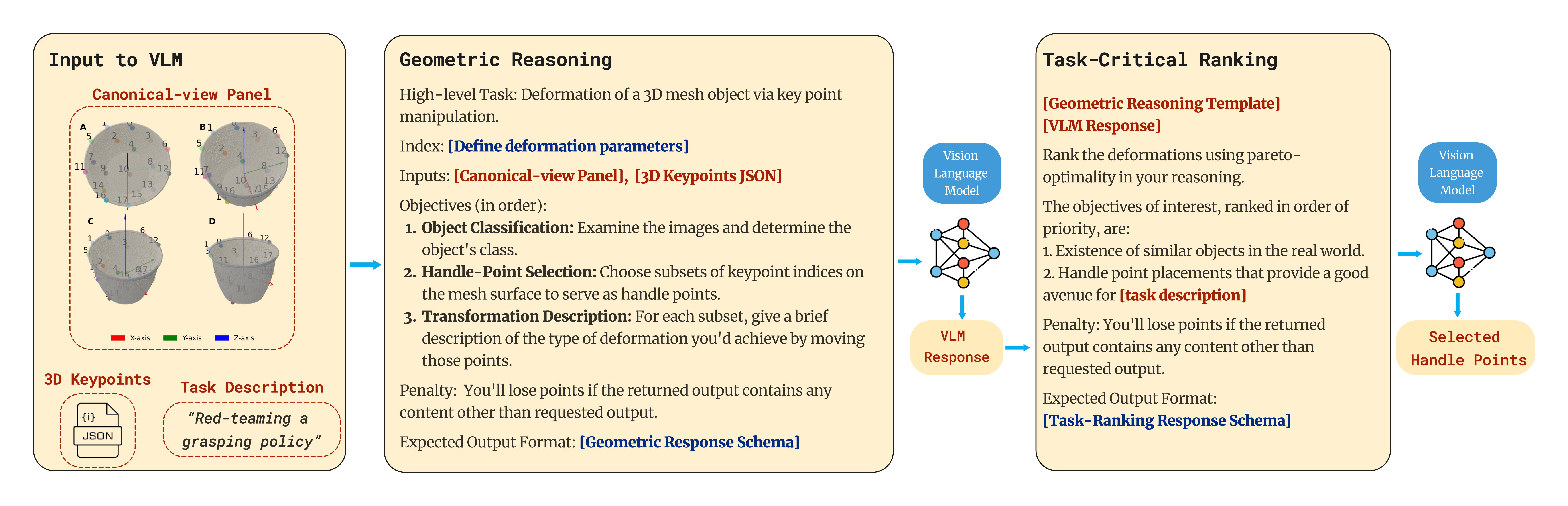}
            \vspace{-7mm}
            \caption{\textbf{Two-stage VLM prompting strategy for 3D handle-point selection.} First, the Geometric Reasoning template aligns a canonical view-panel and indexed keypoints with a high-level task description, guiding the VLM to infer which vertices control meaningful mesh deformations. Next, the Task-Critical Ranking template asks the model to pareto-rank these candidates by plausibility and task relevance, producing a compact set of handle points for targeted, task-aware red-teaming.}
            \label{fig:prompt_template}
            \vspace{-6mm}
        \end{figure}

    \vspace{-2mm}
        We represent each object as a watertight triangular mesh $M = (V, F)$.
        To generate plausible deformations, we specify a set of \textit{handle points} $H \subseteq V$ and \textit{anchor points} $A \subseteq V$ as boundary constraints on the mesh surface, where the former are displaced and the latter remain fixed. 
        We then use the Jacobian-field optimization framework from~\citet{yoo2024plausible} to compute physically coherent perturbations of the nominal mesh.
        Manually specifying $H$ and $A$ at scale is time‐consuming and risks overlooking non‐intuitive failure hypotheses.
        Therefore, we adopt a \emph{VLM-guided} handle selection strategy wherein we prompt a vision-language model ChatGPT-4o to propose candidate handle points based on object geometry and high-level task cues.
        This strategy is realized through a hierarchical prompting framework that guides the VLM through two stages of reasoning (See Figure~\ref{fig:prompt_template}).

        \vspace{-2.5mm}
        \paragraph{Geometric Reasoning Template}
        Given a canonical object mesh $M$, we first generate a \emph{Canonical View Panel}, shown in the left panel of Figure~\ref{fig:prompt_template}---a $2\times2$ grid of rendered views with overlaid, indexed keypoints produced via a heuristic keypoint sampler (see the Appendix~\ref{app:keypoiny_sampler} for details).
        The accompanying prompt instructs the VLM to reason jointly over the visual and 3D coordinate information of the keypoints to propose subsets of handle points, each annotated with semantic descriptions of the intended deformation.

        \vspace{-2.5mm}
        \paragraph{Task-Critical Ranking Template}
        The second prompt asks the VLM to rank these candidate handle point subsets according to a two-part  objective reflecting the dual goals of red-teaming: (i) the plausibility of resulting objects post-deformation, and (ii) their hypothesized potential to induce policy failure.
        The VLM is explicitly instructed to reason in a Pareto-optimal fashion, preferring subsets that achieve a favorable trade-off between structural realism and task-specific red-teaming utility, rather than optimizing for either criterion in isolation (See Figure~\ref{fig:prompt_template}).
        While this approach is generally reliable, we observe failure cases on manufacturing components like USB connectors, likely due to under-representation in the VLM's pre-training corpus.
        In such cases, we allow the user to fall back to manual handle point selection and encode domain knowledge or task-specific priors directly into the deformation process.

    \vspace{-3mm}
    \subsection{Red-Teaming via Black-Box Optimization}
    \label{sec:red_teaming}
    \vspace{-2mm}
        Given a deformation model $D_\theta(M)$ and a black-box manipulation policy $\pi$, we aim to search for deformation parameters $\theta$ that degrade policy performance when evaluated through simulator feedback.
        This problem presents two primary challenges:
        First, the deformation space, defined by displacements of handle points over the object mesh, has several global and local optima.
        Second, the search must be conducted without access to gradients, as neither the reward function nor the simulator are assumed to be differentiable.

        We adopt the selective perturbation strategy from TOPDM~\cite{charlesworth2021solving}, wherein only a random subset of handle-point displacements is perturbed per candidate.
        This design is well suited to high-dimensional deformation spaces, as perturbing all parameters can destabilize globally coherent geometry, while selective updates preserve promising global structure and support fine-grained local refinement.
        In \grtm, this enables incremental adjustment of local geometric features---such as the contour of a contact surface---without introducing large-scale distortions to the object.

        Operationally, our optimization framework maintains a population of deformation candidates sampled from a Gaussian distribution over $\theta$.
        In each iteration, candidates are evaluated via simulator rollouts, ranked according to $\mathcal{J}(\cdot)$, and the highest-performing samples (elites) are used to update the proposal distribution.
        Early iterations apply perturbations broadly, promoting exploration of the global deformation space.
        In later iterations, localized perturbations enable fine-grained adjustment of critical object features that influence policy behavior.
        We implement this optimization loop in NVIDIA IsaacGym~\cite{makoviychuk2021isaac}, evaluating each candidate deformation across a batch of parallel environments with randomized initial poses to obtain a reliable estimate of policy performance.
        The complete procedure for our black-box policy red-teaming framework can be found in Algorithm \ref{alg:optimization}.
        All hyperparameters and task-specific implementation details can be found in Appendix~\ref{app:implementation}.
        
        \begin{center}
        \small
        \scalebox{0.85}{
        \begin{minipage}{\linewidth}
        \begin{algorithm}[H]
        \caption{Red-Teaming Black-Box Manipulation Policies via Simulator Feedback}
        \label{alg:optimization}
        
        \KwIn{
        Task object mesh $M \in \mathbb{R}^3$; Handle points $H = \{h_1, h_2, \dots, h_M\}$, $h_m \in M$; \\
        Anchor points $A \subset M$; \\
        Population size $N$, elite fraction $\rho$, maximum iterations $T$; \\
        Initial Gaussian distribution over deformation parameters: mean $\mu_0 \in \mathbb{R}^{M \times 3}$, diagonal covariance $\Sigma_0 \in \mathbb{R}^{M \times 3}$; \\
        Perturbation fraction $\gamma$ for localized refinement; \\
        Pre-trained manipulation policy $\pi$; \\
        Simulator-based evaluation metric $\mathcal{J}(\pi, D_\theta(M))$.
        }
        
        \KwOut{CrashShape parameters $\theta^* \in \mathbb{R}^{M \times 3}$ inducing minimal $\mathcal{J}(\pi, D_\theta(M))$.}
        
        \vspace{0.2cm}
        \textbf{Initialize:} Sample population $\{\theta_i^{(0)} \}_{i=1}^N$ from $\mathcal{N}(\mu_0, \Sigma_0)$\;
        
        \vspace{0.2cm}
        \For{$t = 1$ to $T$}{
            \ForEach{$\theta_i^{(t-1)}$ in population}{
                Randomly select $\lfloor \gamma M \rfloor$ handle points from $H$\;
                
                Add Gaussian noise to corresponding rows of $\theta_i^{(t-1)}$ drawn from $\mathcal{N}(\mu_0, \Sigma_0)$\;
                
                Generate deformed mesh $\tilde{M}_i = D_{\theta_i^{(t-1)}}(M)$\;
                
                Evaluate task performance: $\mathcal{J}_i = \mathcal{J}(\pi, \tilde{M}_i)$ via simulator rollout\;
            }
            
            Select elite set $\mathcal{E}^{(t-1)}$ = top $\lceil \rho N \rceil$ samples with lowest $\mathcal{J}_i$\;
            
            Replicate elites to form new population $\{\theta_i^{(t)}\}_{i=1}^N$\;
        }
        
        \vspace{0.2cm}
        \Return{$\theta^* = \arg\min_{\theta_i^{(t)}} \mathcal{J}(\pi, D_{\theta_i^{(t)}}(M))$ across all $i, t$.}
        
        \end{algorithm}
        \end{minipage}%
        }
        \end{center}

        

\section{Experiments}
\label{sec:experimental_setup}
\vspace{-2mm}

    We structure our experimental evaluation around three central research questions:
    \vspace{-2.5mm}
    \begin{description}[leftmargin=10pt]
        \item[RQ--1] \emph{Failure Discovery.} Can our red-teaming pipeline reliably uncover catastrophic policy failures through minimal geometric perturbations of nominal objects?
        \vspace{-1mm}
        \item[RQ--2] \emph{Component Contribution.} How much do VLM-guided handle selection and gradient-free optimization individually contribute to the efficiency and quality of failure discovery?
        \vspace{-1mm}
        \item[RQ--3] \emph{Actionability.} Do the generated \emph{CrashShapes} transfer to real-world settings, and can they be leveraged to enhance policy robustness through straightforward fine-tuning?
    \end{description}

    \vspace{-4mm}
    \paragraph{Policies and Object Suites.} 
    We evaluate \grtm~across three robotic manipulation tasks---rigid object grasping, high-precision insertion, and articulated manipulation---to cover diverse policies and failure modes.
    The grasping experiments red-team Contact-Graspnet~\cite{sundermeyer2021contact}, a generalized grasp predictor, on 22 YCB dataset objects~\cite{calli2015benchmarking} that achieve at least 25\% success on nominal shapes under our evaluation protocol (64 grasp trials per object in randomized poses).
    For insertion, we test two variants---a state-based policy trained under the IndustReal framework~\cite{tang2023industreal}, alongside a point-cloud initialized variant using PointNet++~\cite{qi2017pointnet++}. 
    Articulated manipulation employs a state-based drawer-opening policy trained on assets from PartNet-Mobility~\cite{xiang2020sapien}.
    Real-world validation is performed on both the state-based insertion policy and the rigid-object grasping policy (Contact-Graspnet), confirming transferability of discovered CrashShapes beyond simulation.

    \vspace{-3mm}
    \paragraph{Evaluation Metrics.}
    We quantify the effectiveness of our red-teaming framework using four complementary metrics. 
    \textbf{Final Drop} measures the mean relative reduction in success rate from nominal to discovered CrashShapes. \textbf{Iter @ 50\%} indicates the average iterations at which a 50\% relative performance drop is reached.
    \textbf{AUC} is the area under the curve of relative success drop versus iteration, capturing both speed and severity.
    We additionally report the median increase in angular-deficit entropy~\cite{page2003shape})---termed \textbf{$\Delta$Complexity}---relative to the nominal mesh, computed over the ten worst-performing discovered shapes to characterize typical deformation severity while remaining robust to outliers.
    Formal definitions appear in Appendix~\ref{app:evaluation}.

    \vspace{-2.5mm}
    \paragraph{Catastrophic Failure Discovery (RQ--1)}
    Table~\ref{tab:main-results} demonstrates that \grtm~reliably exposes catastrophic failures beyond standard evaluations. 
    In grasping, VLM-guided handles reach severe policy degradation in fewer iterations and without increased geometric deviation compared to manual selection, reflecting the ability of vision–language priors to pinpoint high-leverage contact regions that humans may overlook. 
    For insertion, manual handles outperform VLM proposals on the state-based controller by focusing perturbations on mechanical contact features that drive state-only feedback, while VLM guidance more effectively stresses the point-cloud model by perturbing visually discriminative geometry. 
    In articulated manipulation, manual handles induce near-complete failure immediately. 
    These results demonstrate that the effort required to elicit failure varies with both policy modality and task demands, reinforcing the generality of our approach.
    We additionally report evaluations under a Smoothness Score (SS) constraint, in which we constrain the average handle displacement to be under a threshold.
    Across all tasks, our method continues to induce significant failures in this constrained setting, indicating that the framework does not rely on large deformations to find confounding geometries.
    See the Appendix~\ref{app:smoothness-score} for details.

    \begin{minipage}[!t]{\linewidth}
      \centering
      \small
      \captionof{table}{Red-teaming results across tasks. Final drop, iteration to failure, and AUC measure failure severity; $\Delta$Comp. quantifies geometric deviation.}
      \label{tab:main-results}
      \renewcommand{\arraystretch}{1.2}
      \setlength{\tabcolsep}{7pt}
      \scalebox{0.85}{
      \begin{tabular}{lccccc}
        \toprule
        \textbf{Task} & \textbf{Method} & \textbf{Final Drop (\%)~$\uparrow$} & \textbf{Iter @ 50\%~$\downarrow$} & \textbf{AUC~$\uparrow$} & \boldmath$\Delta$\textbf{Comp.~$\downarrow$} \\
        \midrule
        \multirow{2}{*}{Grasp (YCB)} 
            & VLM-Guided   & 76.3  & 7.3   & 5.26 & 0.041 \\
            & Manual & 63.4 & 9.1 & 4.33 & 0.050 \\
            & VLM-Guided + SS   & 58.3  & 9.2   & 3.301 & 0.002 \\
        \midrule
        \multirow{2}{*}{Articulated Manip.} 
            & VLM-Guided   & 61.9 & 10.0 & 4.90 & 1.517 \\
            & Manual & 98.9  & 6.0   & 6.52 & 0.054 \\
            & Manual + SS & 44.7  & 10.0   & 1.97 & 0.021 \\
        \midrule
        \multirow{2}{*}{Insertion (State)} 
            & VLM-Guided   & 67.4 & 9.0 & 5.53 & 0.286 \\
            & Manual & 73.95 & 8.0   & 5.39 & 0.096 \\
            & Manual + SS & 60.9 & 6.0   & 4.37 & 0.032 \\
        \midrule
        \multirow{2}{*}{Insertion (PCD)} 
            & VLM-Guided   & 77.7 & 6.0 & 6.85 & 0.358 \\
            & Manual & 71.7  & 10.0  & 5.98 & 0.155 \\
            & Manual + SS & 43.4  & 10.0  & 3.85 & 0.044 \\
        \bottomrule
      \end{tabular}
      }
    \end{minipage}

    \paragraph{Ablation Study: Handle Selection and Optimization (RQ--2)}
    We ablate key components of \grtm~by red-teaming Contact-GraspNet~\cite{sundermeyer2021contact} on 22 YCB objects with diverse shapes and grasping affordances. 
    This setting enables controlled comparisons between deformation strategies while retaining object-level variability.
    As shown in Table \ref{tab:ablations}, we factor our pipeline across two key axes---sampling strategy (Gaussian Perturbation vs. Optimization) and handle‐selection (Heuristic vs. VLM-guided)---yielding four core variants. 
    All methods begin by extracting a candidate set of handle points using a fixed geometric heuristic. 
    \textbf{VLM-guided} variants select handles using the prompting strategy from Section~\ref{sec:vlm_mesh_deformation}; while the \textbf{Heuristic} variants uniformly sample handles from the candidate pool, with the selection cardinality matched to VLM mean across all YCB objects for fairness.
    Gaussian perturbation perturbs deformation parameters without structure, while optimization uses TOPDM's selective perturbation scheme. 
    We also include an \textbf{Optimization + All Candidates} baseline that treats all candidates as active handles, increasing both the expressivity and complexity of the search space.
    The results highlight two key findings: (1) optimization substantially improves both failure severity and convergence speed over gaussian perturbations, and (2) VLM-guided handle selection outperforms heuristics, validating the value of learned priors for efficient failure discovery.
    
    \begin{minipage}[!t]{\linewidth}
\centering
\captionof{table}{Ablation results on grasping with Contact-GraspNet across 22 YCB objects. We evaluate the impact of \textbf{handle selection strategy} (Heuristic vs. VLM-guided) and \textbf{deformation search method} (Gaussian Perturbation vs. Optimization). All keypoint-based methods (except “All Handles”) use a fixed handle count matched to the VLM-guided mean. Results show that both VLM guidance and optimization improve failure severity and convergence.}

\label{tab:ablations}
\renewcommand{\arraystretch}{1.2}
\setlength{\tabcolsep}{5pt}
\scalebox{0.85}{
  \begin{tabular}{lcccc}
    \toprule
    \textbf{Method} & \textbf{Final Drop (\%)~$\uparrow$} & \boldmath$\Delta$\textbf{Complexity~$\downarrow$} & \textbf{Iter @ 50\%~$\downarrow$} & \textbf{AUC~$\uparrow$} \\
    \midrule
    Heuristic + Gaussian Perturbation        & 63.3  & 0.058 & 10.00 & 3.654 \\
    Heuristic + Optimization          & 68.4  & 0.035 & 8.95  & 4.610 \\
    All Handles + Optimization        & 71.4  & 0.179 & 8.91  & 4.650 \\
    VLM-Guided + Gaussian Perturbation        & 65.1  & \textbf{0.030} & 10.00 & 3.803 \\
    \midrule
    VLM-Guided + Optimization (Ours)  & \textbf{76.3}  & 0.041 & \textbf{7.32} & \textbf{5.259} \\
    \bottomrule
  \end{tabular}
}

\end{minipage}

    \paragraph{Blue-Teaming: CrashShapes as Corrective Training Signals (RQ--3)}
To assess whether CrashShapes can serve as effective corrective training signals, we fine-tune both insertion policies on subsets of failure-inducing geometries. 
For the state-based policy, we identify two distinct CrashShapes (CS-1, CS-2) and fine-tune separate policy instances on each, alongside the nominal plug. 
Fine-tuning is conducted via PPO with early stopping and no task augmentation. For the point-cloud-initialized policy, we fine-tune a single policy jointly on five CrashShapes (PC-CS-1 to 5) and the nominal plug. 
Table~\ref{tab:blue_sim} shows that across both setups, blue-teaming lifts task success on CrashShapes from 20-45\% to 80–95\%, while preserving original performance on the nominal geometry. 
These results demonstrate that even simple red-teamed geometries can meaningfully guide robustness improvement without inducing regression.

\begin{minipage}[!t]{\linewidth}
\centering
\captionof{table}{
Simulation blue-teaming results on high-precision industrial insertion.
CrashShape performance is reported before and after fine-tuning; the final column confirms nominal performance is preserved.
Nominal pre-training success: 96\% (State-based) and 86\% (PointCloud-initialized).
}
\vspace{-1mm}
\label{tab:blue_sim}
\scalebox{0.85}{
\begin{tabular}{llccc}
\toprule
\textbf{Policy} & \textbf{Geometry} & \textbf{Orig.\,\%} & \textbf{Blue.\,\%} & \textbf{Nominal\,after\,\%} \\
\midrule
\multirow{2}{*}{State-based} & CS-1 & 25.0 & 87.8 & 87.5 \\
                             & CS-2 & 45.0 & 93.8 & 96.0 \\
\midrule
PointCloud-initialized & PC-CS Shapes & 31.3 & 81.3 & 87.3 \\
\bottomrule
\end{tabular}}
\vspace{-2mm}
\end{minipage}

\paragraph{Actionability: Real-World Validation (RQ--3)}
We further validate the practical transferability of red- and blue-teaming to the real-world by fabricating CrashShapes for both the state-based insertion policy and the rigid-object grasping policy (Contact-Graspnet) and evaluating them on hardware. 
Using PLA prints of CS-1 and CS-2, generated via CoACD decomposition~\cite{wei2022approximate} and 3D printing, we conduct 40 physical trials per shape on an xARM 6. 
Results in Table~\ref{tab:blue_real} show a close match to simulation: task success drops from 90\% (nominal) to 22.5\% and 55\% on the CrashShapes, respectively. 
For grasping, we 3D-printed one CrashShape per object for two YCB objects (mustard bottle, screwdriver) and evaluated each in 20 trials on a Franka arm.
In Table~\ref{tab:blue_real}, these appear under ``CS-1'' with ``CS-2'' marked as ``--'', to maintain a uniform column structure across tasks.
The results show substantial real-world success drops, consistent with simulation trends.
Deploying the blue-teamed policies from simulation---without additional real-world fine-tuning---recovers success rates to 90\% (CS-1) and 82.5\% (CS-2), with no degradation on the nominal plug. 
These results affirm the physical realism, and underscore the utility of CrashShapes as both diagnostic and corrective tools. 


\begin{minipage}[!t]{\linewidth}
\centering
\captionof{table}{Real-world validation across insertion and grasping. Columns are uniform for both tasks. 
For insertion, CS-1 and CS-2 are the two printed CrashShapes. 
For grasping, each object has a single printed CrashShape reported under CS-1; CS-2 is “–”. }

\label{tab:blue_real}
\scalebox{0.85}{
\begin{tabular}{l l c c c}
\toprule
\textbf{Task} & \textbf{Policy / Object} & \textbf{Nominal} & \textbf{CS-1} & \textbf{CS-2} \\
\midrule
\multirow{3}{*}{\textbf{Insertion (xARM 6)}}
 & Original Policy        & 90.0\,\% & 22.5\,\% & 55.0\,\% \\
 & Blue-Teaming on CS1    & 85.0\,\% & 90.0\,\% & --       \\
 & Blue-Teaming on CS2    & 95.0\,\% & --       & 82.5\,\% \\
\midrule
\multirow{2}{*}{\textbf{Grasping (Franka)}}
 & Mustard Bottle         & 80.0\,\% & 30.0\,\% & --       \\
 & Screwdriver            & 90.0\,\% & 35.0\,\% & --       \\
\bottomrule
\end{tabular}}
\end{minipage}

\section{Conclusion}
\label{sec:conclusion}

    This paper presents \grtm, an automated red-teaming framework for robotic manipulation policies, with a focus on generating confounding object geometries leveraging user-specified or VLM-guided constraints.
    Our method casts shape deformation as a black-box search problem, using embodied simulation rollouts to discover \textit{CrashShapes}---physically plausible object variants that trigger catastrophic failures in pre-trained manipulation policies.
    The ability to guide deformations using either manual or VLM-derived priors enables semantically grounded stress-testing.
    Importantly, we demonstrate that the failure-inducing geometries discovered in simulation reliably transfer to the real world across multiple manipulation skills, including high-precision insertion on a physical xARM~6 setup and rigid-object grasping on a Franka arm. 
    These CrashShapes can then be leveraged for targeted policy improvement through naive fine-tuning using PPO with early stopping (blue-teaming). 
    This simple training strategy recovers up to 60 percentage points in task success on the CrashShapes without degrading performance on the original shape.
    These results affirm that CrashShapes are not only diagnostic but also actionable, providing a practical pathway to enhance robustness without overfitting.
    Altogether, our pipeline offers a scalable, policy-agnostic tool for structured robustness evaluation and targeted correction of failure modes in robotic manipulation, with validated impact both in simulation and the real world.


\section{Limitations}
\label{sec:limitations}
In this work, we assume that input objects are represented as watertight, manifold triangle meshes.
This requirement arises from the underlying Jacobian field-based deformation model, which depends on well-defined differential operators over the mesh surface.
Real-world scans, however, often contain noise, holes, or non-manifold artifacts, and require careful preprocessing for mesh repair or mesh reconstruction before they can be used within our framework.

Failure-inducing geometries are discovered through embodied simulation rollouts in the Isaac Gym simulator.
Although Isaac Gym offers high-fidelity rigid-body simulation, it inevitably approximates real-world contact dynamics and frictional effects.
While our real-world experiments demonstrate strong transferability for high-precision, millimeter-level tolerance tasks like USB insertion, transferability cannot be assumed universally across all tasks and object types.

Finally, while our framework effectively uncovers hidden failure modes, it does not aim to explain why those specific failures occur. 
The resulting CrashShapes serve as actionable test cases, but interpreting their causal relationship to policy behavior currently requires manual analysis.
Extending the framework with tools for automatic failure diagnosis or causal attribution remains an important direction for future work.


\clearpage
\acknowledgments{We gratefully acknowledge the people and organizations who made this work possible. 
This material is based upon work supported by NIST under Grant No.\ 70NANB24H314. 
This material is also based upon work supported by ONR MURI N00014-24-1-2748. 
We especially thank Zilin Si and Sarvesh Patil for their help with 3D printing the objects used in our real-world experiments. 
We are grateful to the members of the RCHI and R-PAD labs at Carnegie Mellon University for thoughtful feedback that improved the clarity and quality of this paper.}

\subsubsection*{NIST Disclaimer}
Certain commercial entities, equipment, or materials may be identified in this document in order to illustrate a point or concept. 
Such identification is not intended to imply recommendation or endorsement by NIST, nor is it intended to imply that the entities, materials, or equipment are necessarily the best available for the purpose.


\bibliography{citations}  

\clearpage
\newpage
\appendix
\addcontentsline{toc}{section}{Appendix} 
\part{Appendix}
\parttoc 

\section{Qualitative Evolution of Red-Teaming}
\label{app:qual_results}

\begin{figure}[!t]
\centering
\includegraphics[width=0.95\linewidth]{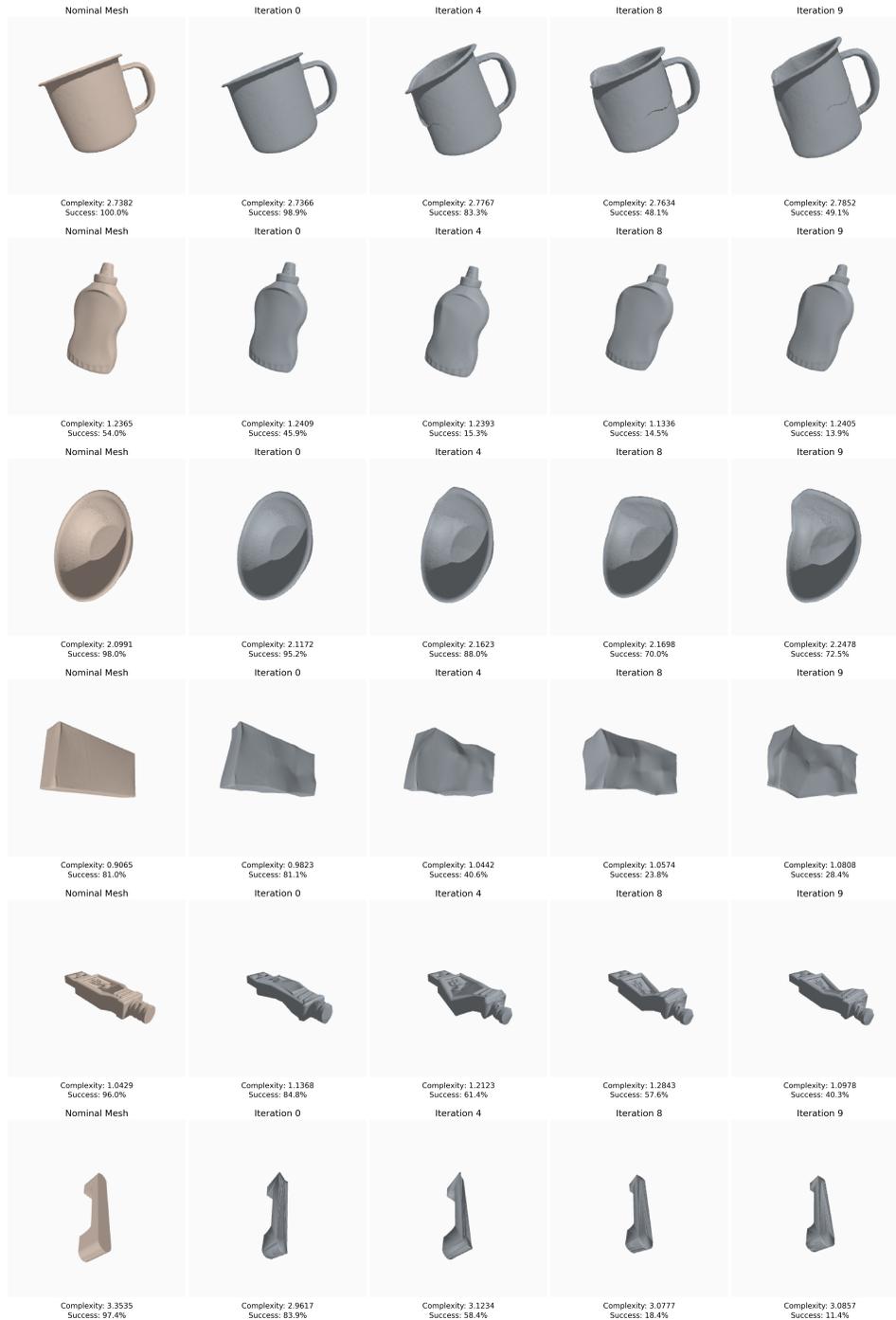}
\caption{\textbf{Evolution of geometric red-teaming across optimization.}
Each row shows an object undergoing deformation via our pipeline across three tasks: rigid grasping (rows 1–4), high-precision insertion (row 5), and articulated drawer manipulation (row 6). Columns show deformation stages with annotated \textbf{shape complexity} and \textbf{task success}.
Results confirm that minor, plausible deformations can collapse performance, often without significant increase in complexity.}
\label{fig:qualitative-evolution}
\end{figure}

To visualize the progression of our geometric red-teaming framework, Figure~\ref{fig:qualitative-evolution} presents the optimization trajectory for six representative objects spanning all three task domains. 
Each row corresponds to a single object, shown at five key stages: the nominal mesh, the initialization (Iteration 0), and Iterations 4, 8, and 9 of the optimization process. 
Annotations include task success rate (measured via simulator rollouts) and morphological shape complexity (computed via angular-deficit entropy).

The first four rows depict grasping objects from the YCB benchmark, while the fifth and sixth rows showcase results from the high-precision insertion and articulated drawer manipulation tasks, respectively. 
These examples reveal that failure-inducing deformations are often subtle: for several objects, catastrophic policy collapse occurs with minimal increase in shape complexity, highlighting the brittleness of learned affordances.

Across tasks, performance degradation is often non-monotonic, reinforcing the need for simulator-in-the-loop evaluation over one-shot or gradient-based strategies. 
Notably, the final CrashShapes in row 5 (USB insertion) and row 6 (drawer manipulation) retain structural realism despite causing near-complete task failure---validating our goal of discovering minimal, plausible perturbations with high diagnostic value.

\section{Evaluating the Role of the 2D Diffusion Prior in APAP}
\label{app:apap_prior}
    Our deformation module adopts only the first stage of the APAP framework~\cite{yoo2024plausible}, which optimizes a per-face Jacobian field subject to handle and anchor constraints. 
    The original APAP method includes a second stage that applies a 2D diffusion prior over mesh texture space to further smooth the deformation field. 
    While beneficial for stylized mesh editing, we find this stage to be unnecessary for our use case, where the primary objective is to produce plausible geometric changes that satisfy localized constraints.

    \begin{figure}[h]
    \centering
    \includegraphics[width=0.9\linewidth]{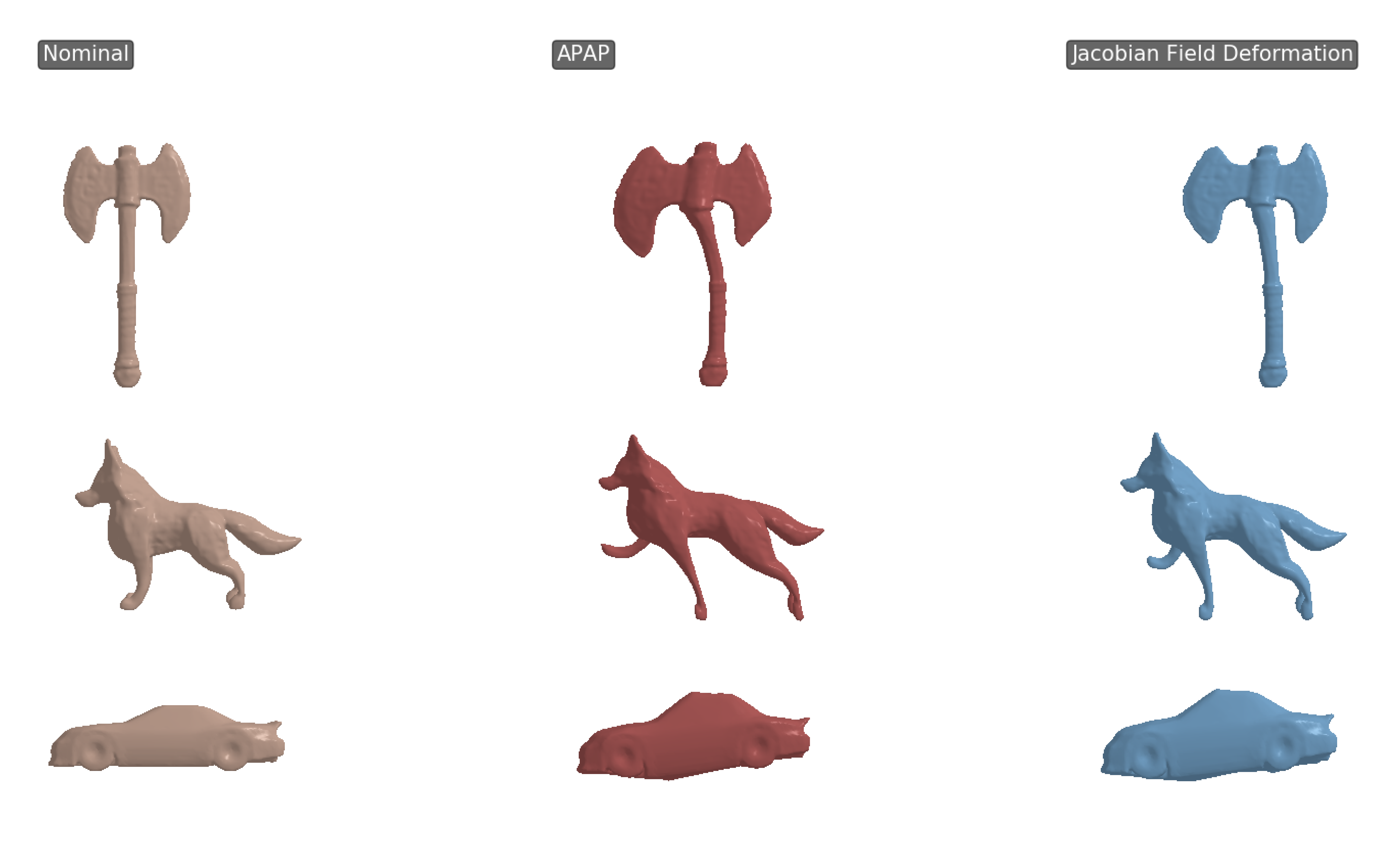}
    \caption{Qualitative comparison of deformations with and without the 2D diffusion prior from APAP. Both variants satisfy identical handle and anchor constraints; differences are minimal despite the Jacobian-only variant being substantially more efficient.}
    \label{fig:apap_ablation}
    \end{figure}

    \paragraph{Deformation Consistency.}
    To evaluate whether omitting the 2D diffusion prior degrades the quality of deformation, we compare the resulting meshes across 15 diverse objects from the original APAP dataset. 
    Figure~\ref{fig:apap_ablation} presents qualitative examples, showing that both deformation pipelines yield smooth, semantically coherent shapes that satisfy the same sets of handle and anchor constraints. 
    Visually, the deformations are nearly indistinguishable, suggesting that the core structure of the transformation is retained without the diffusion stage.
    
    \paragraph{Chamfer Distance.}
    Quantitatively, we compute the pairwise Chamfer Distance between:
    
        the nominal mesh and the full APAP deformation (CD$_\text{Nom-APAP}$),
    
        the nominal mesh and the Jacobian-only deformation (CD$_\text{Nom-Jac}$),
    
        and the two deformed variants themselves (CD$_\text{APAP-Jac}$):
    
    \begin{center}
    \begin{tabular}{lccc}
    \toprule
    & CD$_\text{Nom-APAP}$ & CD$_\text{Nom-Jac}$ & CD$_\text{APAP-Jac}$ \\
    \midrule 
    Value (×1000) & 4.137 & 3.084 & 0.686 \\
    \bottomrule
    \end{tabular}
    \end{center}
    
    The deformation variants differ by an average of just $6.86 \times 10^{-4}$ in absolute terms---remarkably low for objects normalized to a unit bounding box---indicating that the mesh produced by Jacobian optimization alone closely matches the result of the full APAP pipeline.
    
    \paragraph{Computational Efficiency.}
    While accuracy is preserved, the computational footprint differs substantially. 
    On an NVIDIA RTX 4090 GPU, the full APAP pipeline consumes roughly 10 GB of memory and requires 10 minutes per object. 
    In contrast, the Jacobian-only variant completes in 22 seconds using $\sim$1 GB of memory---a 27× speedup in runtime and 10× reduction in memory usage. 
    Given that each optimization run in our red-teaming framework requires hundreds of deformation evaluations, omitting the diffusion prior enables tractable, high-throughput exploration without compromising geometric fidelity.

\begin{figure}[h]
    \centering
    \includegraphics[width=0.9\linewidth]{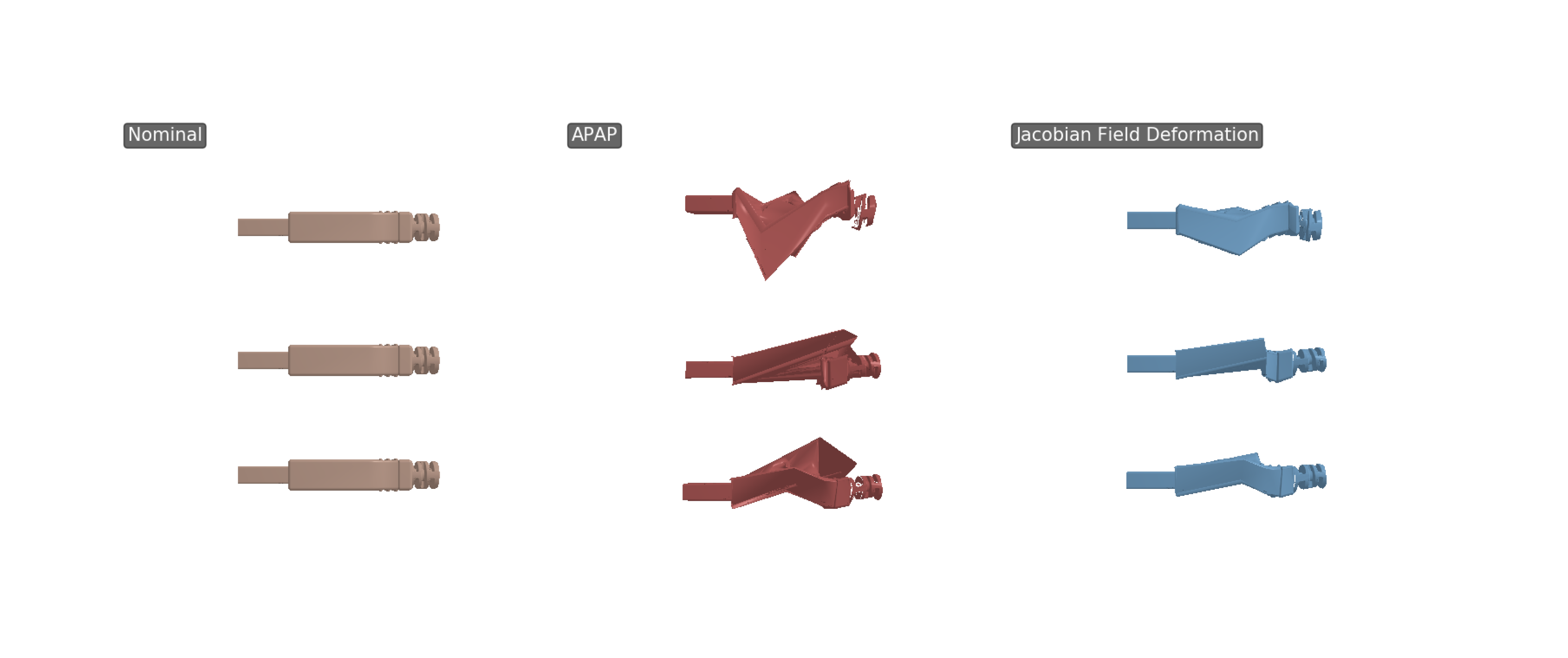}
    \caption{Deformation failure induced by the APAP diffusion prior on a USB plug. While the Jacobian-only variant preserves connector geometry, the full pipeline produces unrealistic deformations. These deviations significantly undermine task viability for insertion.}
    \label{fig:apap_usb}
    \end{figure}

    \paragraph{Failure on Specialized Objects.}
    While the diffusion stage offers negligible benefit in general, it is actively harmful in certain task settings. 
    Figure~\ref{fig:apap_usb} visualizes deformations of a USB plug---an object category likely underrepresented in APAP's 2D training distribution. 
    Here, the full pipeline yields grossly implausible shapes that fail to preserve the structural priors essential to downstream manipulation.

    We quantify this failure by measuring Chamfer Distance in two ways: globally across the full object, and locally over a densely sampled patch on the plug's connector head---the region most critical for the insertion task. 
    Importantly, this region was explicitly constrained via a dense set of anchor points, intended to preserve its geometry during deformation. 
    Despite these constraints, the full APAP pipeline introduces large deviations in this area, indicating that the diffusion prior overrides local geometric fidelity in pursuit of global smoothness. 
    In contrast, our Jacobian-field-only model faithfully preserves the anchor-constrained region while enabling expressive variations elsewhere. 
    The results are summarized below:

    \begin{center}
    \begin{tabular}{lccc}
    \toprule
    Region & CD$_\text{Nom-APAP}$ & CD$_\text{Nom-Jacobian}$ & CD$_\text{APAP-Jacobian}$ \\
    \midrule
    Global (×1000) & 5.074 & 0.793 & 2.429 \\
    Connector Head Only (×1000) & 11.833 & 0.186 & 5.697 \\
    \bottomrule
    \end{tabular}
    \end{center}

    The full APAP pipeline introduces an order-of-magnitude larger deviation in the connector head region than the Jacobian-only variant. 
    This geometric corruption is especially detrimental in tasks like insertion, where small perturbations in contact geometry can lead to complete task failure. 
    In contrast, our Jacobian-field-only model produces constraint-faithful deformations that preserve task-relevant geometry while allowing expressive variations elsewhere.
    
    \paragraph{Takeaway.}
    Omitting the APAP diffusion prior improves deformation throughput by an order of magnitude and yields better geometric preservation on task-relevant regions---especially for specialized objects underrepresented in the prior's training data. 
    Given the absence of meaningful degradation and the significant performance benefits, our framework adopts the Jacobian-only variant for all experiments.

\section{Smoothness Score: A User Interface for Guaranteed Deformation Extents}
\label{app:smoothness-score}

\subsection{Motivation and Design Goals}
We aim to provide a simple interface that gives users \emph{guarantees} on the extent of deformation while preserving mesh plausibility and task semantics. 
Rather than targeting ``visual subtlety,'' which is subjective, we expose a metric, simulator-agnostic \emph{deformation budget} that:
\begin{enumerate}[leftmargin=12pt,itemsep=2pt,topsep=2pt]
    \item provides explicit guarantees on average handle displacement in metric units,
    \item constricts the optimizer to a \emph{narrower feasible set} around the nominal geometry,
    \item composes cleanly with the Jacobian-field solver and anchor constraints from the main method (\sect{sec:method}),
    \item remains stable in optimization via a projection step that preserves proposed directions.
\end{enumerate}

\subsection{Definition and Guarantee}
Let $H=\{h_i\}_{i=1}^{M}$ be the set of handle vertices and $d_i\in\mathbb{R}^3$ their displacements in a candidate deformation. We define the \emph{Smoothness Score} (SS) as
\begin{equation}
\mathrm{SS}(D)\;=\;\frac{1}{M}\sum_{i=1}^{M}\left\lVert d_i\right\rVert_2,\qquad D=\{d_i\}_{i=1}^{M}.
\end{equation}
Given a user budget $\tau>0$ (meters), we collect all budget-feasible deformations in
\begin{equation}
\mathcal{C}_\tau(M)\;=\;\bigl\{\,\theta\in\Theta:\ \mathrm{SS}\bigl(D(\theta)\bigr)\le \tau\,\bigr\}.
\end{equation}
Any proposed candidate with $\mathrm{SS}(D)>\tau$ is projected back to the budget surface by uniform scaling:
\begin{equation}
P_\tau(D)\;=\;s\,D,\qquad s=\min\!\left\{1,\;\frac{\tau}{\mathrm{SS}(D)}\right\}.
\end{equation}
This projection guarantees $\mathrm{SS}\!\left(P_\tau(D)\right)\le \tau$ and preserves the \emph{direction} of the proposed handle motions.

\paragraph{Normalization.}
We report $\tau$ in metric units and also provide axis-aligned bounding-box extents for context. 
A dimensionless variant $\hat{\tau}=\tau/B_{\max}$, where $B_{\max}$ is the maximum box extent, may be used for cross-object comparability.

\subsection{Relationship to Plausibility and $\Delta$Complexity}
The budget limits the \emph{magnitude} of handle motion. 
Plausibility of the full mesh deformation continues to be enforced by the Jacobian-field optimization and anchor constraints. 
We report the median change in angular-deficit entropy (\emph{$\Delta$Complexity}) as a mesh-level proxy for geometric deviation. 
Under budgeted search we observe a significant drop in median $\Delta$Complexity across all task and policy suites, indicating minimal deviation beyond local neighborhoods of the handles.

\subsection{Integration into the Optimization Loop}
We insert the projection after mutation and before the mesh solve. 
This keeps the proposal distribution unchanged while narrowing the feasible region.
\begin{algorithm}[H]
\caption{Budgeted candidate projection within the red-teaming loop}
\label{alg:budget-projection}
\KwIn{Proposed handle displacements $D=\{d_i\}_{i=1}^{M}$, budget $\tau>0$}
\If{$\mathrm{SS}(D)>\tau$}{
    $D \leftarrow \dfrac{\tau}{\mathrm{SS}(D)}\,D$ \tcp*{Uniform scaling to the budget surface}
}
\Return{$D$}
\end{algorithm}

\section{Real-World Setup}
\label{app:real-world}

    \subsection{Insertion Policy}
    Real-world experiments were conducted using an xArm 6 robotic arm connected via Ethernet to a laptop equipped with an NVIDIA GeForce RTX 2070S GPU. 
    Inference and control logic were executed locally on this machine. 
    The xArm was controlled using the manufacturer-provided Python API for Cartesian position control.

    Each CrashShape geometry was processed using CoACD~\cite{wei2022approximate} to generate a convex decomposition. 
    This served two purposes: first, to match the simulation setup, where convex decomposition was used to prevent contact buffer overflows; and second, to improve printability.
    Without decomposition, several CrashShapes exhibited topological artifacts that prevented reliable slicing, such as open holes or thin surfaces; CoACD regularization mitigated these defects.

    \begin{figure}[h]
    \centering
    \includegraphics[width=0.9\linewidth]{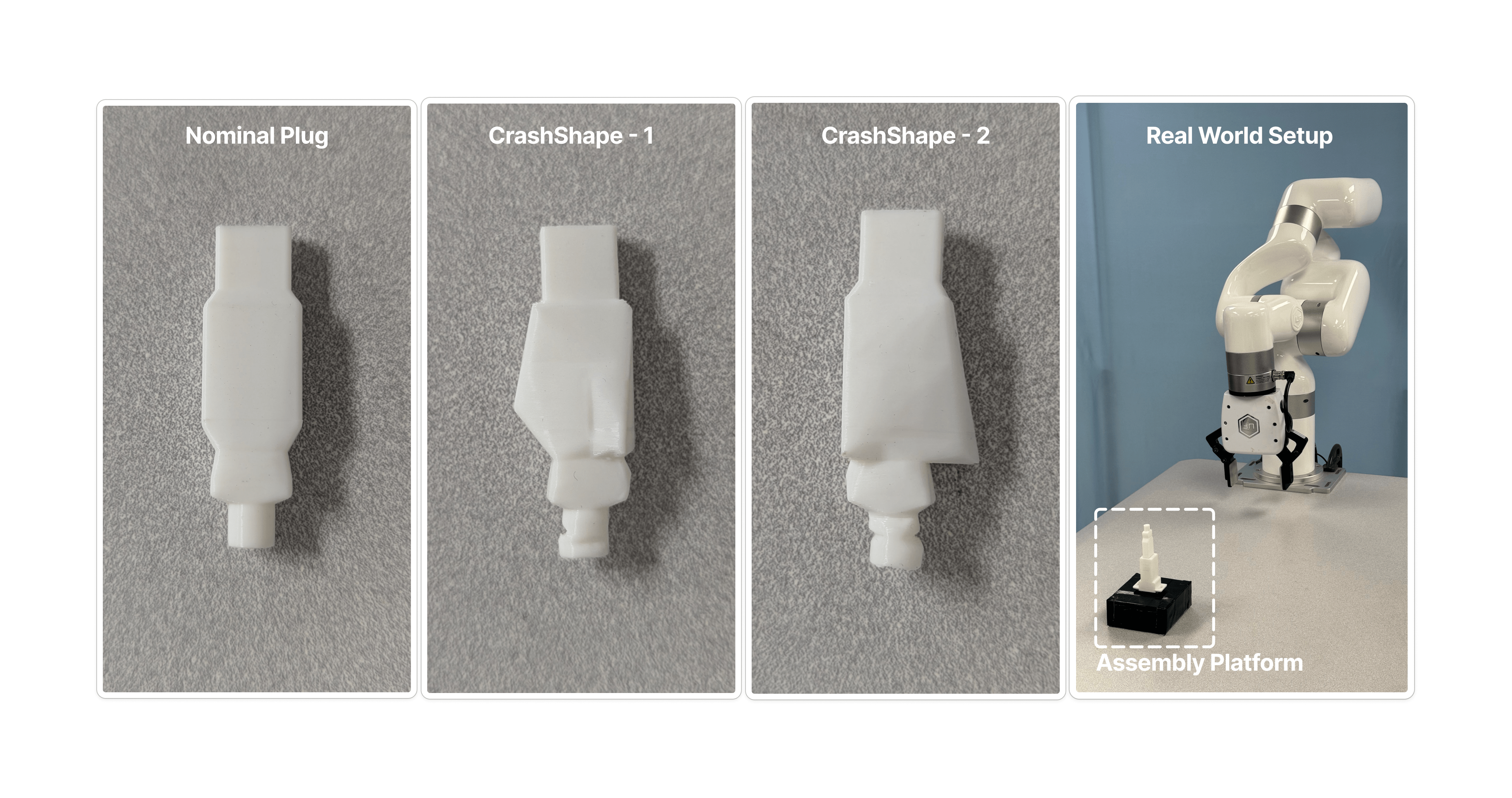}
    \vspace{-5px}
    \caption{\textbf{Physical setup and fabricated geometries used for real-world insertion experiments.} Left: Nominal USB plug and two red-teamed CrashShapes generated by our framework. These 3D-printed variants retain connector plausibility while introducing subtle geometric deviations. Right: xArm 6 robot and assembly platform used for physical testing.}
    \label{fig:real_setup}
    \end{figure}

    The processed meshes were fabricated using a Bambu Labs X1E 3D printer with white PLA filament. Physical socket placement was manually configured to approximate the same relative pose used in simulation. 
    Representative images of the fabricated plugs and socket, as well as the overall experimental setup, are shown in Figure~\ref{fig:real_setup}.
    For control, we used the PLAI controller from IndustReal~\cite{tang2023industreal}, which facilitated policy transfer from Isaac Gym by stabilizing Cartesian-space motions. 
    To adapt the policy outputs to real-world dynamics, a fixed scaling factor was applied to all action vectors prior to execution.

\subsubsection*{Experimental Setup}

Each CrashShape was evaluated across 40 independent insertion trials. 
To ensure geometric diversity and improve coverage of failure modes, we randomized the pose of the plug relative to the socket in each trial. 
Candidate plug poses were sampled using a Poisson Disk distribution centered around the nominal insertion pose and constrained to a bounded cuboidal volume that mirrors the plug initialization distribution used during curriculum-based policy training in simulation. 
This sampling strategy ensured non-overlapping plug placements while maximizing uniformity across the test space, and helped assess policy robustness to small real-world perturbations in plug alignment.

    \subsection{Grasp Policy}
    \label{app:real-world-grasp}
    Real-world grasping experiments were conducted on a table-top Franka Emika Panda arm. An Azure Kinect camera, mounted on a stand, provided RGB-D observations from which object point clouds were extracted. Control was implemented with the Deoxys library using a joint position controller: given a target grasp pose from the policy, we solved inverse kinematics for the target joint configuration and commanded the arm to that configuration.
    
\begin{figure}[h]
    \centering
    \includegraphics[width=0.95\linewidth]{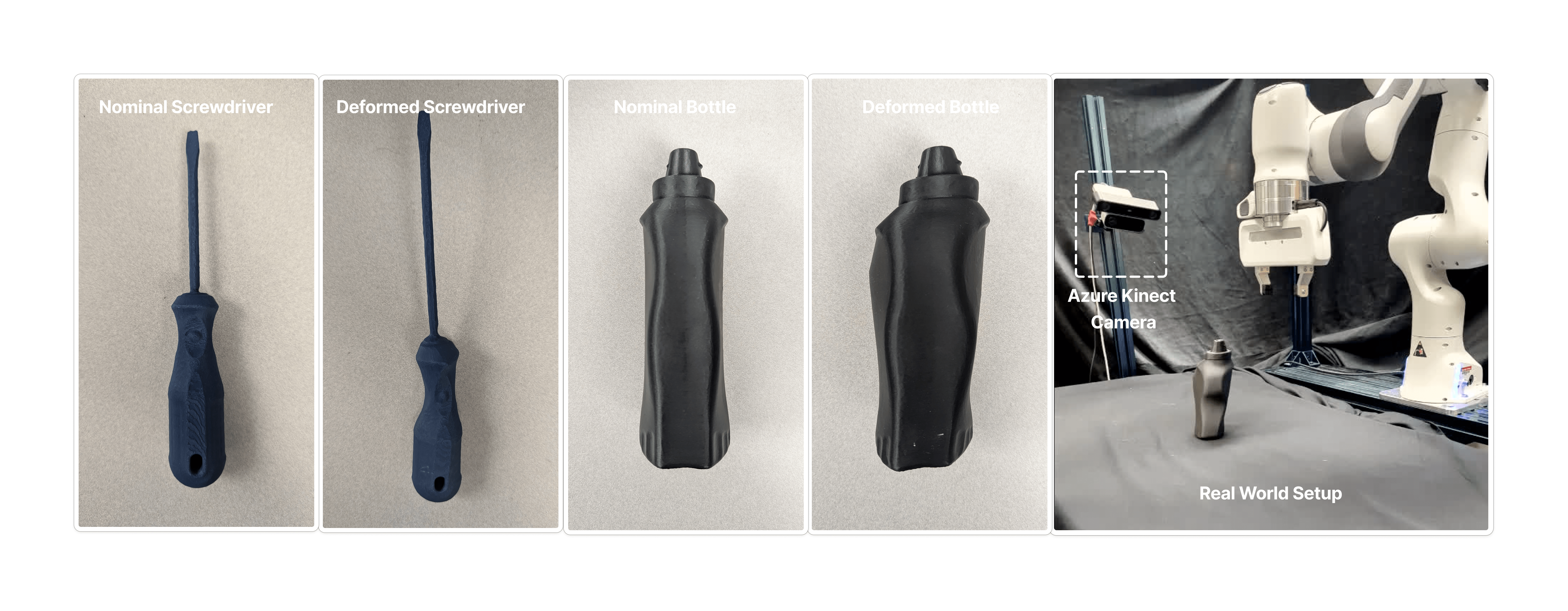}
    \vspace{-5px}
    \caption{\textbf{Physical setup and fabricated geometries used for real-world grasping experiments.} 
    Left: Nominal screwdriver and bottle and their CrashShapes (deformed variants). 
    Right: Table-top Franka Emika Panda with an Azure Kinect camera used to acquire point clouds.
    These 3D-printed variants preserve plausibility while altering geometry relevant to grasping.}
    \label{fig:grasp_real_setup}
    \end{figure}

    Nominal and GRT-generated CrashShape geometries were 3D printed in PLA (Bambu Labs X1E). The prints preserve object plausibility while introducing subtle geometric deviations that affect grasp stability. Representative objects and the physical setup are shown in Fig.~\ref{fig:grasp_real_setup}.
    
    \subsubsection*{Experimental Setup}
    Each object (nominal and CrashShapes) was evaluated over 20 independent grasp trials. For each trial, the object was randomly re-initialized on the tabletop within a bounded workspace region with randomized planar position and yaw. The grasp policy proposed a target pose from the point cloud; the pose was executed via the IK-then-joint-position pipeline described above.


\section{Evaluation Metrics}
\label{app:evaluation}

    We evaluate the effectiveness of our red-teaming framework using four metrics: relative final drop in success, area under the degradation curve (AUC), mean iterations to first catastrophic failure, and median change in shape complexity. 
    These metrics quantify failure severity, optimization efficiency, and geometric plausibility.
    All metrics are computed per object, then aggregated across the benchmark suite. 
    Formal definitions and justifications follow.

    \paragraph{Final Drop in Success Rate.}
    Let $S_0$ denote the success rate of a pre-trained policy $\pi$ on the undeformed object $M$. For each red-teamed geometry $T_\theta(M)$ discovered during optimization, let $S\theta$ denote the success rate of $\pi$ when evaluated on that geometry.
    The final drop is defined as the maximum relative decline in performance:
    \begin{equation*}
        \Delta\text{final} \coloneqq \text{max}_{\theta \in \Theta} (\frac{S_0 - S_\theta}{S_0}),
    \end{equation*}

    where $\Theta$ is the set of all deformation parameters explored by the red-teaming framework.
    
    \emph{Motivation.} This metric captures the worst-case relative degradation in policy performance across the entire optimization trajectory. 
    By normalizing to the nominal object geometry, it supports consistent comparison across tasks with differing baseline success rates.

    \paragraph{Area Under Degradation Curve (AUC).}
    Let $S_t$ denote the best success rate observed up to iteration $t$, and define the normalized degradation curve as:
    \begin{equation*}
    C(t) \coloneqq (\frac{S_0 - S_t}{S_0}),t=0,\cdots,T.    
    \end{equation*}

    We compute AUC via trapezoidal integration over the $T$ optimization steps:
    \begin{equation*}
        \text{AUC} \coloneqq \sum_{t=1}^{T} \frac{1}{2} (C(t) + C(t-1)).
    \end{equation*}
    \emph{Motivation.} AUC jointly reflects the rate and severity of failure discovery. 
    High AUC corresponds to early and substantial degradation, making it a sensitive indicator of optimizer effectiveness under constrained simulation budgets.

    \paragraph{Mean Iterations to First Catastrophic Failure.}
    A geometry $T_\theta(M)$ is classified as inducing catastrophic failure if:
    $S_\theta \leq 0.5 \cdot S_0$.
    Let $t_{\text{fail}}^{(i)}$ be the first optimization step at which this condition is met for object $i$. Then,
    \begin{equation*}
        \text{E}[t_{fail}] \coloneqq \frac{1}{N}\sum_{i=1}^{N}t_{fail}(i),
    \end{equation*}
    where $N$ is the number of test objects.
    \emph{Motivation.} This metric quantifies how efficiently the optimizer surfaces critical failure cases---those that reduce policy performance by at least 50\% relative to the undeformed object. 
    It provides insight into the convergence dynamics of the search process.

    \paragraph{Median Change in Shape Complexity.}
    We adopt the angular-deficit entropy~\cite{page2003shape} as a morphological proxy for shape complexity. For a mesh $M$, the angular deficit at vertex $j$ is given by:
    \begin{equation*}
        \Phi_j = 2\pi - \sum_i\phi_i,
    \end{equation*}
    where $\phi_i$ are the internal angles of faces adjacent to $j$. 
    The histogram of $\Phi_j$ over $[-2\pi, 2\pi)$ is normalized into a probability distribution $p(\Phi_b)$ over bins $b$, and the entropy of this distribution defines the complexity:
    \begin{equation*}
        H(M) \coloneqq - \sum_b p(\Phi_b)\log p(\Phi_b).
    \end{equation*}

    We compute this measure for the nominal object geometry and for the ten red-teamed geometries with lowest observed success rate. 
    The reported metric is the median increase in complexity:
    \begin{equation*}
        \Delta \text{Complexity} \coloneqq \text{median}_{\theta \in \Theta \prime}(H(T_\theta(M)) - H(M)),
    \end{equation*}
    where $\Theta' \subset \Theta$ is the set of ten worst-performing geometries.
    \emph{Motivation.} This metric estimates the typical morphological deviation required to induce failure. 
    We use the median rather than the mean or maximum to suppress the influence of extreme outliers and better characterize the central tendency of deformation severity.
    Prior work has shown that angular-deficit entropy correlates well with human intuition for shape complexity, making it a meaningful descriptor for plausibility.


\section{Canonical View Panel Construction}
\label{app:canonical_views}

    To support 3D keypoint selection via the \textit{Geometric Reasoning} prompt, we construct a 4-view canonical panel that depicts a mesh overlaid with semantically diverse surface keypoints. This section details the in-house heuristic sampling algorithm used to select these keypoints, as well as the rendering strategy used to generate the final composite.
    
    \subsection{Keypoint Sampling via PCA-Guided Geometric Heuristics}
    \label{app:keypoiny_sampler}
    Given a watertight object mesh $M$, we first uniformly sample $P = 20{,}000$ surface points $\{x_i\}_{i=1}^P$ and their corresponding face normals. A two-stage downsampling procedure is then applied to promote semantic and spatial diversity while preserving geometric symmetry and coverage.
    
    \paragraph{Stage 1: Principal Axis Estimation.}  
    We compute the mean-centered point cloud $X \in \mathbb{R}^{P \times 3}$ and extract its dominant principal direction $v_1 \in \mathbb{R}^3$ using Principal Component Analysis (PCA):
    \[
    v_1 := \arg\max_{\|v\| = 1} \text{Var}(X v),
    \]
    with subsequent orthogonal directions $\{v_2, v_3\}$ forming a right-handed basis. These axes define a canonical frame for symmetry-aware sampling.
    
    \paragraph{Stage 2: Symmetry-Aware Candidate Selection.}  
    We begin by estimating a local density $\rho(x_i)$ at each point $x_i$ using $k$-nearest neighbor distances:
    \[
    \rho(x_i) := \left( \frac{1}{k} \sum_{j=1}^k \|x_i - x_{j(i)}\| \right)^{-1},
    \]
    where $\{x_{j(i)}\}_{j=1}^k$ are the $k$ nearest neighbors of $x_i$. Points are ranked by $\rho$ to prioritize sampling in regions with structural detail.
    
    For each high-density candidate point $x_i$, we compute its reflection about the principal axes:
    \[
    x_i^{(j)} := x_i - 2 (v_j^\top x_i) v_j, \quad j \in \{1,2,3\},
    \]
    and identify the nearest mesh point $x^* \in \{x_k\}_{k=1}^P$ to each reflection using a KD-tree query. Reflected candidates are accepted only if they satisfy a minimum distance constraint $\|x^* - x'\|_1 > \delta$ for all previously selected keypoints $x'$, ensuring spatial separation.
    
    This stage yields approximately half the desired keypoints, promoting symmetric, well-distributed coverage.
    
    \paragraph{Stage 3: Axis-Aligned Completion.}  
    The remaining samples are drawn from projections of the point cloud onto each principal axis. Points are sorted by projected position along $v_j$, and farthest-point sampling is performed within this ordering to ensure geometric spread while avoiding oversampling along minor features.
    
    \paragraph{Output.}  
    The final set of $N = 25$ keypoints is obtained by combining symmetric and axis-aligned samples, followed by a final pruning pass that filters pairs of points with Euclidean separation less than $\delta = 0.1$.
    
    \subsection{Canonical View Rendering and Panel Assembly}
    To construct the canonical panel, we render the mesh from ten viewpoints sampled uniformly over the viewing hemisphere using spherical coordinates:
    \[
    (\text{azimuth}, \text{elevation})_i = \left( 180 \cdot \frac{i}{n}, \, 180 \cdot \left(0.5 - \frac{i+0.5}{n} \right) \right), \quad i = 0, \dots, n{-}1,
    \]
    where $n = 10$. From these, four diverse views are selected and rendered using a fixed 3D perspective.
    
    Each rendered view overlays the mesh $M$ with the selected keypoints. Keypoints are visualized using a perceptually uniform colormap with one color per index, enabling direct cross-view correspondence. Numerical indices are overlaid at each point to facilitate unambiguous referencing by vision-language models.
    
    The object mesh is rendered with low-opacity shading to ensure that keypoint markers and annotations remain visible, even in regions of geometric occlusion or low curvature. The global coordinate frame is plotted at the mesh origin with axes colored in RGB (X: red, Y: green, Z: blue), providing a consistent spatial reference across objects and views.
    
    The resulting four images are arranged into a $2 \times 2$ grid with quadrant labels (A--D) to form a single composite image used as input to the \textit{Geometric Reasoning} prompt.
    

\section{VLM Prompts}
\label{app:vlm_prompt_templates}

    \subsection{Geometric Reasoning Prompt}
    To ground handle point selection in semantically meaningful and physically plausible regions of the mesh, we initiate our hierarchical prompting pipeline with a stage we term \textit{Geometric Reasoning}. 
    This stage is designed to interface 3D mesh representations with the inductive biases of image-language foundation models, which are typically pre-trained on paired natural language captions and perspective renderings of real-world scenes, and therefore lack native support for reasoning over metric 3D geometry.

    Our input consists of two complementary components. 
    First, a panel of rendered canonical views with overlaid keypoints provides spatial context through image-space structure and surface topology. 
    Second, a serialized point cloud exposes the underlying 3D coordinates of these keypoints in the world frame, enabling the model to localize correspondences across views and infer geometric relationships. 
    Together, these inputs guide the model toward identifying deformation-relevant subsets of keypoints that are grounded in both perceptual semantics and physical plausibility.

    \subsubsection*{Prompt Design and Structured Output Schema}
    
    The prompt consists of two primary inputs:
    \begin{itemize}
        \item A set of \textbf{canonical views}, showing the object from multiple angles with overlaid keypoints.
        \item A \textbf{3D keypoint point cloud}, represented as a JSON file listing coordinates and vertex indices.
    \end{itemize}

    These inputs are paired with an instruction block that defines the deformation task, introduces the concepts of handle and anchor points, and specifies the expected output format. 
    The goal is to elicit sets of keypoints that can be displaced to yield meaningful shape transformations while preserving structural plausibility.

    To ensure that the output from the VLM is machine-parseable and semantically rich, we enforce a structured response schema defined using Python's \texttt{pydantic} interface. Each response includes:
    \begin{itemize}
        \item \texttt{keypoint\_indices}: indices of mesh vertices to be used as handle points.
        \item \texttt{semantic\_object\_label}: a brief description of the object part affected.
        \item \texttt{expected\_transformations}: a list of plausible geometric operations that the selected keypoints could enable.
    \end{itemize}

    The schema is defined as follows:
    \begin{lstlisting}[language=python,caption={Structured schema for geometric reasoning responses.},label={lst:response_schema},basicstyle=\footnotesize\ttfamily]
    class HandlePoint(BaseModel):
        keypoint_indices: List[int]
        semantic_object_label: str
        expected_transformations: List[str]
    
    class Choices(BaseModel):
        choices: List[HandlePoint]
    \end{lstlisting}

This design encourages both high-level conceptual grounding and fine-grained geometric localization. 
The returned handle sets are later filtered and ranked in the second stage of our pipeline.

\subsubsection*{Concrete Example: Bowl Object Prompt}

We now provide the full text of a prompt issued to the model for the \textit{bowl} object from the YCB dataset. 
This example combines the 3D keypoint data with the deformation task description in a format that is directly interpretable by a VLM.

\begin{lstlisting}[caption={Geometric Reasoning prompt issued to the VLM for the bowl object.},label={lst:bowl_prompt},basicstyle=\footnotesize\ttfamily]
High-level Task: Deformation of a 3D mesh object via key point manipulation.

Index:

1. Handle Points -- Mesh vertices which serve as keypoints to be displaced for the purposes of mesh deformation.

2. Anchor Points -- Mesh vertices which remain in place through the deformation process.

3. Displacement Vectors -- Directional displacements of handle points for the purposes of mesh deformation.

Method: Jacobian field optimization by means of a Poisson solver over the entire mesh under the constraints imposed by the deformation parameters.

Input:

1. Images containing multiple canonical views of the object under inspection, annotated with keypoints sampled on the object's surface.

2. A JSON file containing the 3D location of the object in the world coordinates.

Objective: Provide multiple subsets of keypoints to serve as handle points for the deformation process, along with a single line description of the transformations one can hope to achieve using those handle points.

Hints: 
1. The multiple annotated views of the underlying object, along with the 3D world frame positions of the keypoints, can be used to localize the keypoints and reason about the type and structure of the object.
2. Deformations that yield realistic objects often maintain symmetry across the key axes of symmetry.

Penalty: 
You'll lose points if the returned output contains any content other than requested output.

Json File:
---
{
    "points": [
        {"index": 0, "coordinates": [-0.47172870409604656, -0.14247249438098228, 0.17822726655889315]},
        {"index": 1, "coordinates": [-0.46929558729234583, 0.1350456377707993, 0.1796215656020141]},
        {"index": 2, "coordinates": [-0.35644749303156864, -0.3464904332560477, 0.17204569944492737]},
        {"index": 3, "coordinates": [-0.35300153906438475, 0.017063087435454195, -0.07459165106927007]},
        {"index": 4, "coordinates": [-0.32169580652204455, -0.1885024930547734, -0.03901991261143628]},
        {"index": 5, "coordinates": [-0.26721085819530344, 0.3453163712585918, 0.0891211484372599]},
        {"index": 6, "coordinates": [-0.20121465343061082, 0.21445315574996734, -0.13554194451645532]},
        {"index": 7, "coordinates": [-0.16289154898156788, -0.46736681905226585, 0.17699411386750605]},
        {"index": 8, "coordinates": [-0.07177334220021714, -0.36468495429770964, -0.0461483319136234]},
        {"index": 9, "coordinates": [-0.04224310529134742, 0.3813957476898451, -0.028733817963148323]},
        {"index": 10, "coordinates": [-0.039131664935924744, -0.1252406967012443, -0.14648472527402323]},
        {"index": 11, "coordinates": [0.0508298903781717, 0.17015377379407495, -0.14869136388269089]},
        {"index": 12, "coordinates": [0.09356827578313848, 0.4913365426332097, 0.16198085363100448]},
        {"index": 13, "coordinates": [0.1902896398639075, -0.37800836951965583, 0.06340997268906587]},
        {"index": 14, "coordinates": [0.20253074338317512, -0.295063278758562, -0.06585252822312912]},
        {"index": 15, "coordinates": [0.2047127648919393, 0.3362792723809077, -0.023198523693772254]},
        {"index": 16, "coordinates": [0.26734059954429806, -0.022174095770825998, -0.14745247326570718]},
        {"index": 17, "coordinates": [0.38576119336327147, 0.22145619287954538, 0.12381330003252163]},
        {"index": 18, "coordinates": [0.3992351923636997, -0.28827945973258284, 0.1774309357560207]},
        {"index": 19, "coordinates": [0.43362864405756985, -0.04558597752419907, 0.09125812443992103]}
    ]
}
---
\end{lstlisting}

This prompt provides the model with both semantic structure and geometric detail, facilitating informed and localized handle selection for downstream mesh deformation.

\subsubsection*{Example Output: Parsed Response from Geometric Reasoning Prompt}

For the prompt issued on the \textit{bowl} object (Listing~\ref{lst:bowl_prompt}), the vision-language model returned the following structured response, automatically parsed according to the schema described in Listing~\ref{lst:response_schema}. 
Each entry identifies a candidate set of handle keypoints, a semantic description of the affected part, and the expected outcome of deformation:

\begin{lstlisting}[caption={Parsed output returned by the VLM for the bowl object, identifying semantically meaningful handle regions.},label={lst:bowl_response},basicstyle=\footnotesize\ttfamily]
{
  "0": {
    "semantic_object_label": "bowl",
    "keypoint_indices": [0, 1, 2, 5, 12],
    "expected_transformations": [
      "Expand or contract the rim of the bowl."
    ]
  },
  "1": {
    "semantic_object_label": "bowl",
    "keypoint_indices": [9, 11, 15, 19],
    "expected_transformations": [
      "Alter the shape of the bowl's side profile."
    ]
  },
  "2": {
    "semantic_object_label": "bowl",
    "keypoint_indices": [6, 18],
    "expected_transformations": [
      "Adjust the height or depth of the bowl."
    ]
  }
}
\end{lstlisting}

This response illustrates the model’s capacity to associate localized subsets of mesh vertices with intuitive part-level semantics and deformation goals, even when reasoning only over projected renderings and sparse 3D coordinates.

\subsection{Task-Critical Ranking Prompt}
The second stage of our hierarchical prompting pipeline refines the handle point proposals generated during geometric reasoning by evaluating their utility with respect to a concrete downstream manipulation task. 
We refer to this stage as \textit{Task-Critical Ranking}.

\subsubsection*{Motivation and Design}

While the first stage encourages the model to propose a diverse set of plausible deformation handles grounded in visual semantics and geometric symmetry, not all suggestions are equally meaningful for the intended manipulation objective (e.g., grasping, insertion). 
To disambiguate these proposals and select the most task-relevant subset, we issue a follow-up query that reframes deformation selection within the broader context of policy stress-testing.

This second-stage prompt is issued in the presence of the full conversational history from the Geometric Reasoning stage, including the canonical view panel, point cloud representation, and the model's prior output. 
By replaying the earlier interaction, we ensure that the ranking query remains referentially grounded and contextually coherent.

Critically, the model is instructed to apply a Pareto-optimality criterion when reasoning about deformation utility. 
Specifically, the prompt enumerates two prioritized objectives: (1) the existence of real-world analogs for the resulting geometry, and (2) the potential of the deformation to challenge or degrade a grasping policy. 
The model is asked to identify candidate subsets that lie on the Pareto frontier of this multi-objective tradeoff---favoring geometries that are both physically plausible and adversarially informative.

The final output is expected to be a single top-ranked deformation candidate, serialized in a strict JSON format conforming to the schema described below.

\subsubsection*{Response Schema}

To ensure consistent and parseable outputs from the model, we enforce a tightly scoped response schema. The expected output is a single top-ranked candidate, encapsulated in a typed JSON object as follows:

\begin{lstlisting}[language=python,caption={Schema used to parse VLM responses for the task-critical ranking stage.},label={lst:ranking_schema},basicstyle=\footnotesize\ttfamily]
class HandlePoint(BaseModel):
    keypoint_indices: List[int]
    semantic_object_label: str
    expected_transformations: List[str]

class TopRank(BaseModel):
    top_choice: HandlePoint
\end{lstlisting}

This schema ensures that the final output consists of a single handle region, distilled from the larger candidate set, and tailored to the semantics and physical structure most likely to affect task performance.

\subsubsection*{Prompt Example}

The full text of the prompt issued to the model at this stage is shown below.
The conversational history from the prior Geometric Reasoning stage is replayed before this message to retain semantic grounding:

\begin{lstlisting}[caption={Raw prompt string issued during the Task-Critical Ranking stage.},label={lst:ranking_prompt},basicstyle=\footnotesize\ttfamily]
Rank the deformations using pareto-optimality in your reasoning. The objectives of interest at the end of the deformation process, ranked in order of priority, are:

1. Existence of similar objects in the real world.
2. Handle point placements that provide a good avenue for red-teaming a grasping policy.

Objective: Return a single json file containing the information on the highest ranking subset.

Penalty: You'll lose points if the returned output contains any content other than requested output.
\end{lstlisting}


\section{Operational Details of the Red-Teaming Framework}
\label{sec:appendix:redteaming_hparams}

Our red-teaming framework employs a conservative, geometry-aware optimization strategy to identify subtle yet task-critical object deformations that induce policy failure. 
All experiments reported in this paper use our adapted version of the TOPDM algorithm~\cite{charlesworth2021solving} to navigate the deformation space, subject to anchor and handle constraints derived either from user specification or automated selection via a hierarchical prompting strategy.

\subsection*{Deformation Space Initialization}

The deformation space is defined over a normalized and centered mesh representation, with vertex coordinates scaled to fit within a unit cube. 
This normalization ensures numerical stability during optimization and compatibility with the Poisson-based deformation model. 
All candidate geometries are rescaled to their original dimensions and aligned to the undeformed shape via an Orthogonal Procrustes transformation using the anchor point correspondences. 
This alignment step is essential to preserve physically meaningful contact regions---particularly for insertion and articulation tasks where positional accuracy governs success.

\subsection*{Perturbation Strategy and Optimization Parameters}

To preserve realism and avoid geometric artifacts, we initialize the deformation search with a low-variance Gaussian distribution over the Jacobian field parameters. 
The standard deviation is fixed to $0.001$ (in normalized mesh units), which strikes a balance between exploration and structural integrity. 
A larger variance was found to produce frequent self-intersections or mesh degeneracies, while a smaller one limited the optimizer's ability to discover non-trivial failure modes. 

The optimizer evaluates a batch of 10 candidates per iteration and runs for a fixed budget of 10 iterations. 
A fractional sampling strategy---central to TOPDM---is used to selectively perturb only half of the parameters at each iteration. 
This mechanism facilitates fine-grained, local search without collapsing global structure, which is particularly important for preserving shape semantics while probing failure boundaries.
In practice, red-teaming a single object--policy pair requires between 0.5 and 4 wall-clock hours on a single NVIDIA RTX~4090, depending on the policy and observation space dimensionality.

\subsection*{Simulation and Evaluation Constraints}
Each deformation candidate is evaluated in a physically realistic simulation environment using task-specific performance metrics computed under a pre-trained manipulation policy. 
These metrics vary by task. 
For both insertion and grasping, we use binary success criteria: insertion success is determined based on positional alignment and contact constraints, while grasping success is defined by the object’s retention following a lift-and-shake test. 
For articulated manipulation, we instead adopt a continuous metric corresponding to the final displacement of the drawer's prismatic joint, allowing the optimizer to reason over graded failure severity rather than discrete outcomes.
In all cases, the optimization objective is minimized to surface deformations that degrade policy effectiveness.

To ensure stable simulation dynamics and compatibility with collision detection routines, each deformed mesh undergoes convex decomposition using CoACD~\cite{wei2022approximate} prior to loading. 
This step mitigates issues such as contact buffer overflows, non-manifold surface artifacts, and slicing errors that may arise during 3D printing or simulation-based evaluation. 
It also mirrors the preprocessing applied to assets during training, helping maintain consistency across nominal and perturbed evaluations.

Grasping and articulation tasks inherit their contact models and solver settings directly from Isaac Gym. 
The deformed geometry is introduced per rollout by replacing the nominal mesh, allowing the policy to interact with physically plausible but strategically perturbed shapes we term \textit{CrashShapes}.


\section{Implementation Details} 
\label{app:implementation}

    \subsection{Simulation Environment Configuration} 
    All simulation experiments were conducted using NVIDIA Isaac Gym Preview Release 4~\cite{makoviychuk2021isaac}, which provides GPU-accelerated physics via the NVIDIA PhysX engine. To exploit parallel rollouts, the workspace was divided into four independent subscenes executed concurrently on a single CUDA-enabled GPU.
    
    The simulation advanced at \SI{60}{\hertz} with two internal substeps per frame, stabilizing contact dynamics and rigid body interactions. We employed the Temporal Gauss-Seidel (TGS) solver with 16 position iterations per substep. Gravity was set to \SI{9.81}{\meter\per\second\squared} along the negative $z$-axis, and global damping was enabled to mitigate oscillatory motion.
    
    All experiments used a Franka Emika Panda manipulator with self-collision checking. A fixed random seed (42) governed environment initialization, and PyTorch’s deterministic mode was disabled to maximize throughput.
    
    Object meshes were incorporated based on task-specific requirements. For grasping and insertion, all deformed shapes (\textit{CrashShapes}) were preprocessed using CoACD~\cite{wei2022approximate} to generate convex rigid-body approximations compatible with the physics engine. The insertion task additionally incorporated contact feedback using signed distance field (SDF) queries to model penetration and alignment. In contrast, articulated manipulation experiments used undecomposed triangle meshes directly.
    
    \subsection{Grasping Task: Environment, Policy, and Evaluation} 
    
    \subsubsection{Environment Setup} 
    The grasping benchmark was instantiated in 64 parallel environments. Objects were drawn from a curated subset of 22 YCB models~\cite{calli2015benchmarking}, each selected based on achieving at least $25\%$ nominal grasp success using the pre-trained policy on unmodified meshes. This baseline success rate was computed using the same perturbation-tolerant evaluation described below, averaged over randomized trials.
    
    At the beginning of each trial, a single object was placed at the center of a planar surface with uniform perturbations applied to its $XY$ position ($\pm\SI{0.02}{\meter}$) and yaw orientation ($\pm\SI{0.79}{\radian}$). The robot was reset to a fixed configuration at episode start.
    
    Perception relied on six simulated depth cameras placed at canonical positions on a virtual hemisphere focused on the object origin. Each camera captured $240 \times 360$ depth images, which were fused to produce a complete point cloud representation of the scene.
    
    \subsubsection{Grasping Policy} 
    We used a pre-trained Contact-GraspNet (CGN)~\cite{sundermeyer2021contact} model as the grasping policy. Depth images were segmented to isolate object points, transformed into CGN’s expected coordinate frame, and processed to generate grasp proposals. If no candidates were detected, inference was repeated up to 10 times.
    
    The highest-ranked valid grasp was selected after geometric consistency checks. This pose was adjusted by a \SI{5}{\centi\meter} approach offset and a \SI{90}{\degree} rotation about the approach axis before execution. The resulting $6$D grasp was converted into joint-space commands using inverse kinematics with a Damped Least Squares (DLS) solver and tracked using Isaac Gym’s built-in PD controller.
    
    \subsubsection{Evaluation Protocol} 
    Each grasp attempt comprised three phases: approach, grasp, and lift. After securing the object, the robot applied a horizontal shake---displacing the gripper \SI{10}{\centi\meter} laterally and returning---to test stability under dynamic motion.
    
    A grasp was deemed unsuccessful if the object slipped from the gripper at any point during or after the shake. We report success rate $J(\pi, M)$ averaged over 64 trials per object, each trial corresponding to one parallel environment with randomized initial conditions.
    
    \subsubsection{Metric Motivation and Red-Teaming Implications} 
    This evaluation protocol goes beyond assessing contact feasibility to test post-grasp stability under external perturbation. Such criteria surface fragilities in grasps that may lift objects reliably yet fail under minor disturbances---an essential distinction for real-world deployment.
    
    By identifying object geometries that degrade this stability metric, our red-teaming framework reveals failure modes tied not to perception alone, but to the physical affordances exploited by the grasping policy. Subtle changes in surface curvature, contact patch topology, or fingertip alignment can result in superficially valid grasps that are dynamically brittle. Discovering such \textit{CrashShapes} offers insight into the geometric assumptions baked into CGN’s grasp predictions and where they begin to fail.
    
    \subsection{High-Precision Insertion Task} 
    
    \subsubsection{Environment Setup} 
    The insertion task was simulated across 128 parallel environments. In each trial, the socket pose was randomized: $XY$ displacements were drawn uniformly from $\pm\SI{0.1}{\meter}$, vertical height was offset in $[\SI{0.0}{}, \SI{0.05}{}]\,\si{\meter}$, and yaw perturbed within $\pm\SI{0.087}{\radian}$. The plug was spawned above the socket using a curriculum-driven offset, with additional $XY$ noise of $\pm\SI{0.01}{\meter}$ when aligned over the rim.
    
    \subsubsection{Policy Architecture and Training} 
    Both the state-based and point cloud-based insertion policies shared the same neural architecture:
    \begin{itemize}
        \item A reduced-size classification-type PointNet++ encoder producing a $32$D geometric feature vector;
        \item A $2$-layer LSTM with 256 units per layer and layer normalization;
        \item A feedforward MLP (512-256-128 units, ELU activation) consuming the concatenated LSTM output and encoder input.
    \end{itemize}
    Policies were trained using PPO with a learning rate of \num{1e-4} (linearly decayed), discount factor $\gamma = 0.998$, clipping threshold of $0.2$, and 8 mini-epochs per update. Rollouts used a horizon of 128 steps and minibatches of size 8192. The reward function combined multiple terms from IndustReal~\cite{tang2023industreal}: SDF-based proximity (sampled at 1000 points), interpenetration checks (SAPU, \SI{1}{\milli\meter} threshold), an engagement bonus, and SBC-based curriculum shaping. Training continued for up to 8192 epochs.
    
    \subsubsection{State-Based Variant} 
    The input to the policy comprised a $24$D vector: 7 joint positions, a $7$D end-effector pose, a $7$D noisy target pose (with up to $\pm\SI{1}{\milli\meter}$ $XY$ noise), and a $3$D displacement from the current end-effector pose to the target. The policy output a $6$D delta pose command, scaled by $0.01$ and applied via a task-space impedance controller.
    
    \subsubsection{Point Cloud Variant} 
    The point cloud-based policy received geometric input from multiple cameras. Depth images were segmented and aggregated into a unified scene point cloud, which was downsampled to 500 points using Farthest Point Sampling. This centered point cloud was passed to the PointNet++ encoder, and its output concatenated with proprioceptive state before being processed by the LSTM and MLP.
    Action generation and control were identical to the state-based variant.
    
    \subsubsection{Evaluation} 
    Each episode terminated after 256 steps or earlier upon success. A trial was considered successful if two conditions were met: the plug's final vertical distance to the socket base was less than \SI{3}{\milli\meter}, and the mean distance between four annotated plug-socket keypoint pairs was below \SI{0.15}{\meter}. Success rates were averaged over 128 independent trials.
    
    \subsection{Articulated Manipulation Task} 
    
    \subsubsection{Environment Setup} 
    This task involved opening drawers attached to cabinet assets with functional prismatic joints, including the sektion cabinet from Isaacgym and models from PartNet-Mobility~\cite{xiang2020sapien}. 
    Each episode began with the drawer fully closed and the robot reset to a perturbed joint configuration with additive uniform noise of $\pm\SI{0.25}{\radian}$.
    
    Simulation parameters included a single substep per frame, 12 position iterations, 1 velocity iteration, a contact offset of \SI{0.005}{\meter}, and a maximum depenetration velocity of \SI{1000}{\meter\per\second}.
    
    \subsubsection{Policy Details} 
    The state-based policy received a $23$D observation vector: 9 joint positions, 9 scaled joint velocities (scaled by $0.1$), a $3$D vector from the end-effector to the drawer handle grasp pose, and the drawer’s current position and velocity ($1$D each). 
    The policy produced $9$D joint deltas (7 arm, 2 gripper DoFs), clipped to $\pm 1.0$, scaled by $7.5$, and applied via the low-level PD controller.
    PPO training rewarded proximity to the grasp frame, end-effector alignment, accurate gripper finger placement, maximum drawer translation, and penalized large control inputs.
    
    \subsubsection{Evaluation} 
    Episodes were capped at 500 steps. The task success metric $J(\pi, M)$ was defined as the final value of the drawer’s prismatic joint position. 
    The theoretical maximum opening was \SI{0.39}{\meter}. Results were averaged over 4096 randomized trials.
    

\end{document}